\RequirePackage{xcolor}
\documentclass[journal,twoside,web]{ieeecolor}

\usepackage{etoolbox}

\makeatletter
\@ifundefined{color@begingroup}%
  {\let\color@begingroup\relax
   \let\color@endgroup\relax}{}%
\def\fix@ieeecolor@hbox#1{%
  \hbox{\color@begingroup#1\color@endgroup}}
\patchcmd\@makecaption{\hbox}{\fix@ieeecolor@hbox}{}{\FAILED}
\patchcmd\@makecaption{\hbox}{\fix@ieeecolor@hbox}{}{\FAILED}
\makeatother

\usepackage{amsmath,amssymb,amsfonts}
\usepackage{bm}
\usepackage{algorithm}
\usepackage{algorithmic}
\usepackage{array}
\usepackage{makecell}
\usepackage{multirow}
\usepackage{multicol}
\usepackage{booktabs}

\def\arraystretch{1.2}

\usepackage{graphicx}
\usepackage{wrapfig}
\usepackage{subfigure}
\usepackage{stfloats}

\usepackage{cite}
\usepackage{textcomp}
\usepackage{url}
\usepackage{verbatim}
\usepackage[normalem]{ulem}
\usepackage{pifont}

\usepackage{CJKutf8}

\usepackage{scalerel}
\usepackage{tikz}
\usetikzlibrary{svg.path}

\definecolor{orcidlogocol}{HTML}{A6CE39}
\tikzset{
    orcidlogo/.pic={
        \fill[orcidlogocol] svg{M256,128c0,70.7-57.3,128-128,128C57.3,256,0,198.7,0,128C0,57.3,57.3,0,128,0C198.7,0,256,57.3,256,128z};
        \fill[white] svg{M86.3,186.2H70.9V79.1h15.4v48.4V186.2z}
        svg{M108.9,79.1h41.6c39.6,0,57,28.3,57,53.6c0,27.5-21.5,53.6-56.8,53.6h-41.8V79.1z M124.3,172.4h24.5c34.9,0,42.9-26.5,42.9-39.7c0-21.5-13.7-39.7-43.7-39.7h-23.7V172.4z}
        svg{M88.7,56.8c0,5.5-4.5,10.1-10.1,10.1c-5.6,0-10.1-4.6-10.1-10.1c0-5.6,4.5-10.1,10.1-10.1C84.2,46.7,88.7,51.3,88.7,56.8z};
    }
}

\newcommand\orcidicon[1]{\href{https://orcid.org/#1}{\mbox{\scalerel*{
                \begin{tikzpicture}[yscale=-1,transform shape]
                \pic{orcidlogo};
                \end{tikzpicture}
}{|}}}}

\usepackage{jsen}

\def\BibTeX{{\rm B\kern-.05em{\sc i\kern-.025em b}\kern-.08em
    T\kern-.1667em\lower.7ex\hbox{E}\kern-.125emX}}

\definecolor{abstractbg}{rgb}{0.89804,0.94510,0.83137}
\definecolor{lilac}{RGB}{192, 141, 255}

\definecolor{limegreen}{RGB}{92, 189, 130}

\newcommand{\etal}{\textit{et al.} }

\newcommand{\notes}[1]{}

\usepackage[colorlinks]{hyperref}

\usepackage{booktabs}
\usepackage{multirow}
\usepackage{graphicx}
\usepackage{xcolor}
\begin{document}
\begin{CJK}{UTF8}{gbsn}

\title{Teacher-Guided Multimodal Dual Representation Reconstruction for Egocentric Human Activity Recognition}

\author{Yuzhe Hao\textsuperscript{\orcidicon{0000-0001-7400-0837}}, 
Ikuro Sato\textsuperscript{\orcidicon{0000-0001-5234-3177}}, \IEEEmembership{Member, IEEE},
Rei Kawakami\textsuperscript{\orcidicon{0000-0003-2342-3324}}, \IEEEmembership{Member, IEEE}, \\
Koichi Shinoda\textsuperscript{\orcidicon{0000-0003-1095-3203}}, \IEEEmembership{Senior Member, IEEE}
}


\IEEEtitleabstractindextext{%
\fcolorbox{abstractbg}{abstractbg}{%
\begin{minipage}{\textwidth}%
\begin{wrapfigure}[13]{c}{3in}
\includegraphics[width=2.8in]{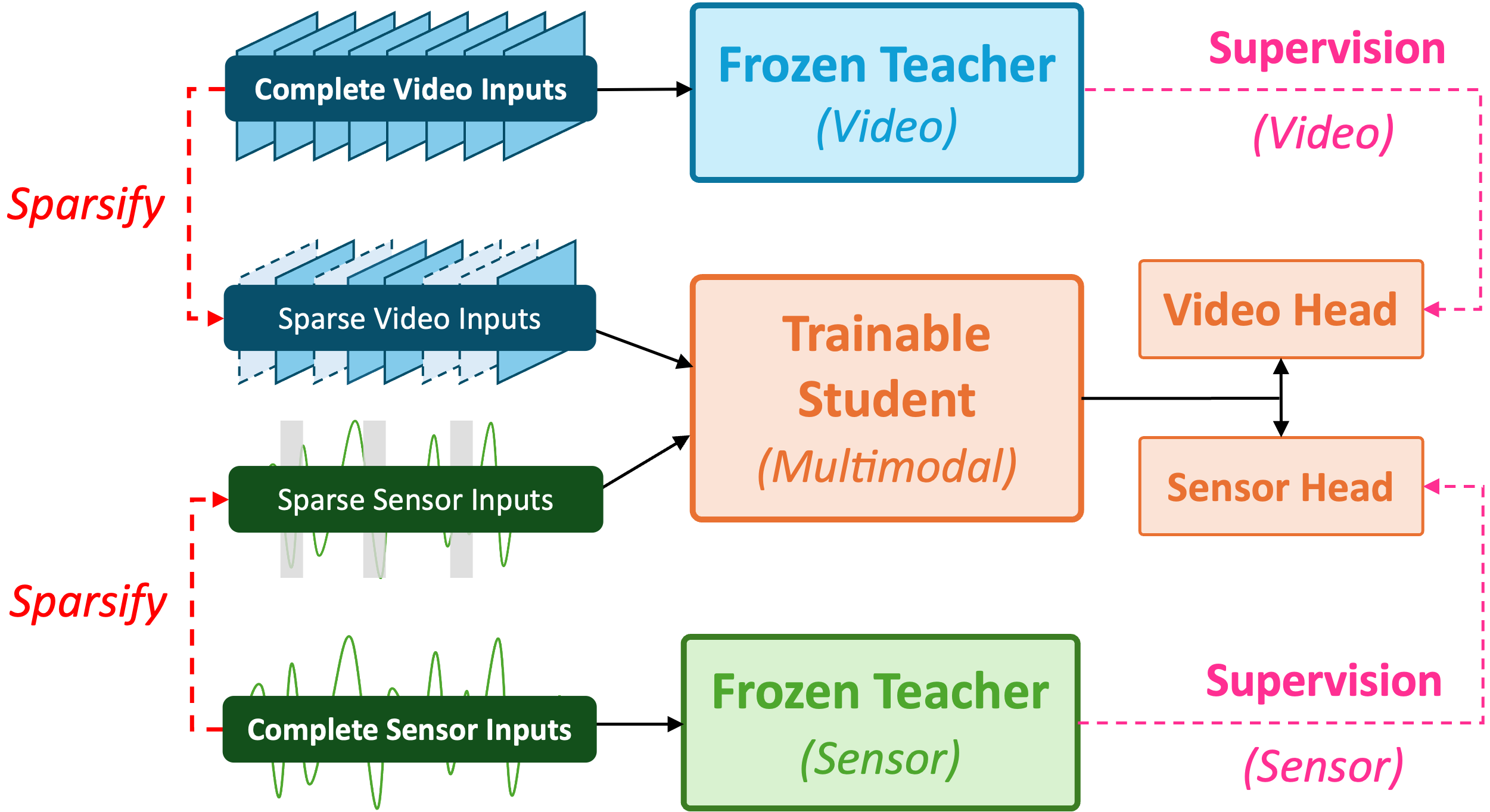}
\end{wrapfigure}
\begin{abstract}
Multimodal Egocentric Human Activity Recognition (Ego-HAR) leverages complementary cues from first-person vision (V) and inertial sensor data (S) to recognize daily activities,
where video mainly captures appearance and object-related context while inertial signals capture motion cues that video may miss.
However, existing pre-training methods typically supervise the fused representation with a single joint objective, which may underemphasize the complementarity between the two modalities and limit the model's ability to learn from distinct modality-oriented signals.
To address this issue, we propose Multimodal Dual Representation Reconstruction (MDR2), a teacher-guided pre-training framework in which frozen, label-informed video and inertial teachers each provide dense, modality-specific targets to a multimodal student trained on temporally sparse inputs.
On the EvIs-Kitchen dataset, MDR2 improves noun/verb F1-macro over the supervised M-Baseline by 9.20/6.17 points at K=4 video key frames and by 1.21/1.82 points at K=32.
These results show that MDR2 consistently achieves higher mean recognition performance across the evaluated video-observation settings under the unified backbone setting, with more pronounced gains under sparse inputs.

\end{abstract}

\thanks{This work is an outcome of a research project, Development of Quality Foundation for Machine-Learning Applications, supported by DENSO IT LAB Recognition and Learning Algorithm Collaborative Research Chair (Science Tokyo). It was also supported by JST CREST JPMJCR1687 and JSPS KAKENHI JP23H00490.
}

\end{minipage}}}

\maketitle


\section{Introduction}

\IEEEPARstart{E}{gocentric} human activity recognition (Ego-HAR) aims to recognize actions from first-person observations.
This setting closely reflects how humans interact with the physical world by capturing visual attention and the objects being engaged with \cite{damen2022rescaling}.
To complement the visual stream, inertial sensor data provide a motion-oriented perspective for activity understanding.
In particular, wrist-worn inertial sensors can continuously and reliably track subtle hand movements, 
and combining them with egocentric video can achieve a more comprehensive and robust understanding of human actions \cite{song2016egocentric}, \cite{song2016multimodal}, \cite{imran2018multimodal}, \cite{hao2024egocentric}.
In this paper, we focus on studying Ego-HAR under this visual-inertial multimodal setting.

To enhance representation learning for Ego-HAR, self-supervised learning (SSL) \cite{chen2020simCLR}, \cite{hsu2021hubert}, \cite{radford2018improving} has emerged as an effective approach for leveraging unlabeled data.
This is particularly important for multimodal Ego-HAR, where fully synchronized multimodal annotations are limited and costly to collect.
Existing multimodal SSL methods mainly follow contrastive or reconstruction-based paradigms. 
Contrastive methods \cite{owens2018audio},
\cite{tsutsui2021you}, \cite{akbari2021vatt} learn representations by aligning paired modalities, while reconstruction-based methods 
\cite{sarkar2023self}, \cite{lian2023av} guide learning by reconstructing input data or latent representations.

In visual-inertial Ego-HAR, effective multimodal learning requires modeling the different roles of egocentric videos and inertial sensors.
Since different modalities encode distinct aspects of an activity \cite{huang2021makes}, egocentric videos mainly provide object-, appearance-, and scene-related cues, whereas inertial sensors provide motion-related cues such as hand movements and action dynamics.
This heterogeneity places two requirements on pre-training supervision: it should distinguish visual- and motion-oriented learning objectives \cite{du2023uni}, and it should remain informative when video observations are sparse or irregular, as is common in practical wearable applications. An effective pre-training framework should therefore provide modality-specific supervision that stays informative under sparse visual input.

\begin{figure}[!t]
\centering{
    \subfigure[Contrastive Siamese Framework \cite{owens2018audio}, which supervises multimodal representations by emphasizing shared components across modalities.]{
        \includegraphics[width=3.3in]{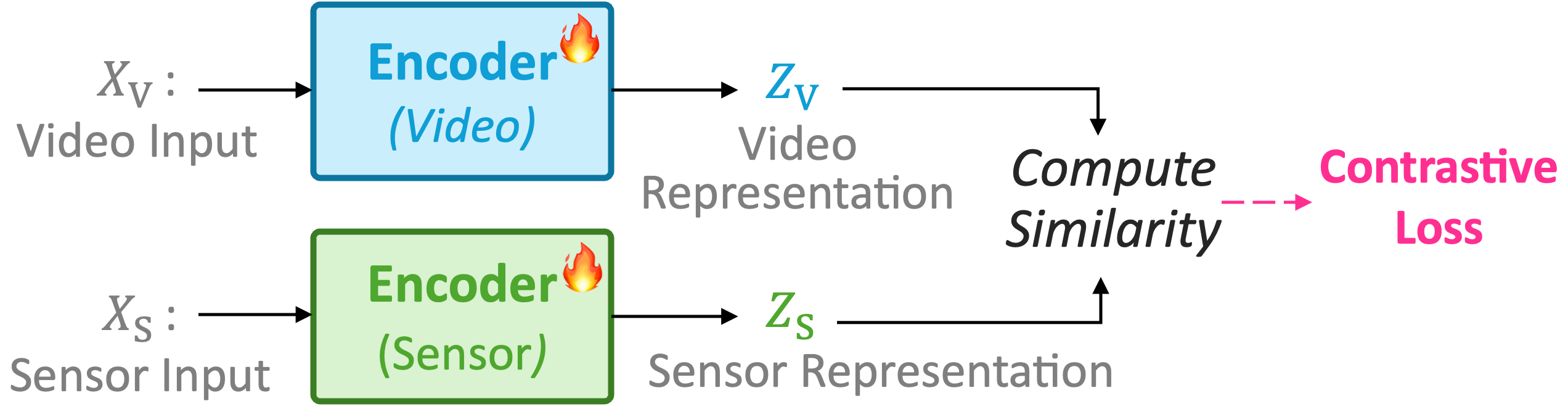}
        \label{fig:ssl-contra}
    }
    \hfil
    \subfigure[Representation Reconstruction Framework \cite{hinton2015distilling}, which uses a single joint reconstruction loss to supervise multiple modalities, without explicitly separating modality-oriented output spaces.]{
        \includegraphics[width=3.3in]{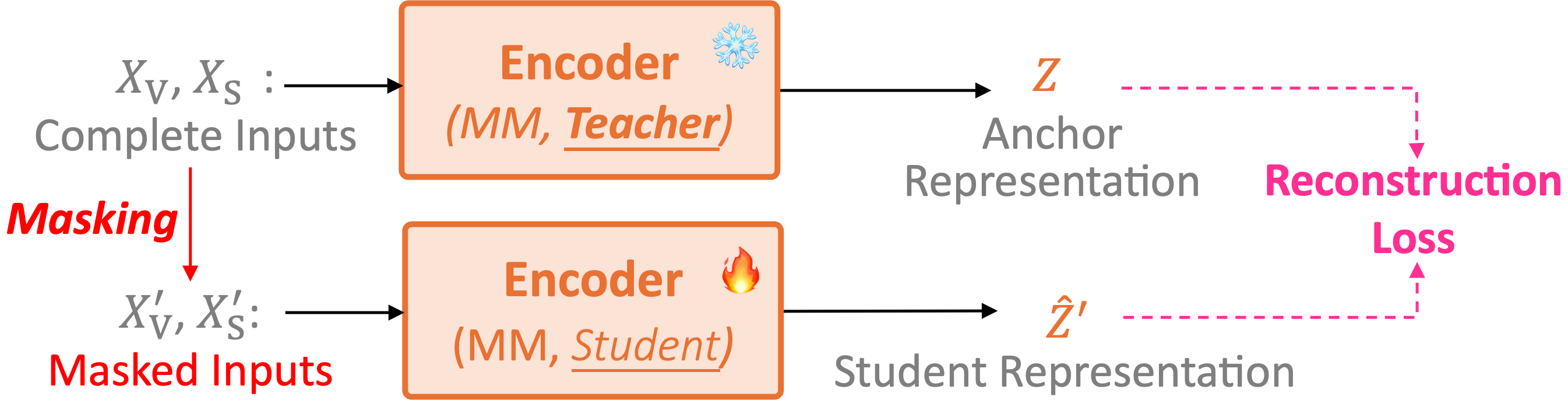}    
        \label{fig:ssl-rep-recon}
    }
    \hfil
    \subfigure[Our Proposed MDR2 Framework, which supervises a fused multimodal representation using dual unimodal reconstruction targets.]{
        \includegraphics[width=3.3in]{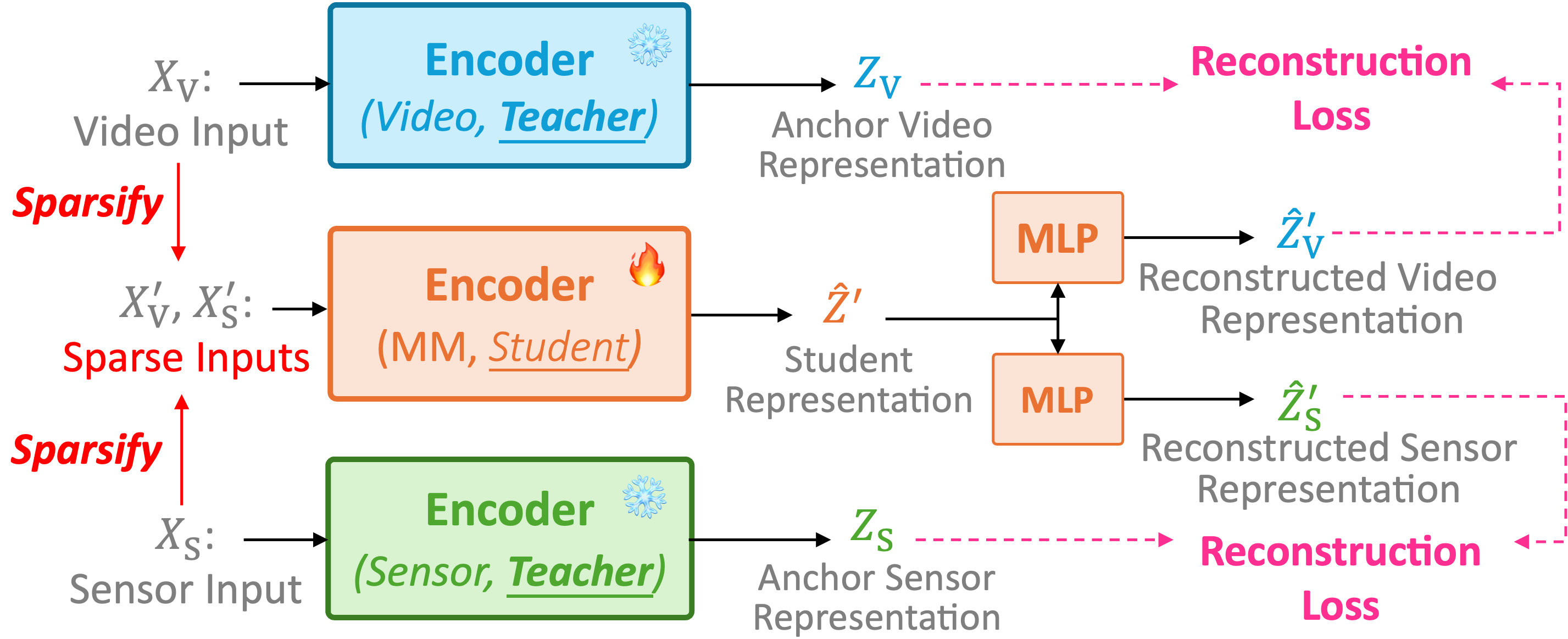}
        \label{fig:ssl-our}
    }
    \caption{Comparing our MDR2 with previous multimodal pre-training frameworks.}
}
\vspace{-0.4cm}

\label{fig:three-ssl}
\end{figure}

Designing such a modality-aware multimodal pre-training framework for visual-inertial Ego-HAR with sparse inputs remains challenging.
In multimodal settings, contrastive pre-training (Fig.~\ref{fig:ssl-contra}) may overemphasize shared information between modalities \cite{dufumier2024align}, \cite{swetha2023presrving}, which can limit its performance on classification and recognition tasks \cite{zhang2024learning}.
Reconstruction-based pre-training (Fig.~\ref{fig:ssl-rep-recon}) avoids relying on augmentation pairs and can provide stable target representations for student models during pre-training \cite{he2022masked, uelwer2025survey}.
However, multimodal variants of this paradigm often rely on a single joint reconstruction target to supervise the fused representation. 
Such unified supervision provides one signal for both modalities without explicitly distinguishing their visual- and motion-oriented learning objectives.

To address this structural limitation of single-target multimodal supervision,
we propose \textbf{M}ultimodal \textbf{D}ual \textbf{R}epresentation \textbf{R}econstruction (MDR2), 
a sparse-to-dense reconstruction pre-training framework that provides modality-specific teacher-target supervision to the fused representation (Fig.~\ref{fig:ssl-our}).
Motivated by practical settings in which modality-specific models or supervision may be more readily
available than fully synchronized multimodal annotations,
MDR2 uses two frozen unimodal teachers to generate dense visual and sensor anchors for supervising a shared multimodal student from temporally sparsified inputs.
The fused representation is trained to reconstruct both anchors through dual unimodal objectives.
The teacher encoders are adapted using activity labels from the training split. Therefore, although the MDR2 reconstruction loss does not directly use activity labels, its reconstruction targets are label-informed.

We evaluate MDR2 on the EvIs-Kitchen dataset \cite{hao2023evis}, which provides temporally synchronized egocentric videos and bilateral wrist inertial sensor data together with verb and noun labels. This combination enables a controlled evaluation of visual- and motion-related recognition performance.
MDR2 achieves higher mean F1-macro performance than the evaluated multimodal baselines and pre-training frameworks under the unified-backbone setting.
We further compare MDR2 with EviMAE~\cite{zhang2025evimae}, a recent masked autoencoding framework based on spectrogram inertial inputs.
In particular, under the limited visual-input setting with only four video key frames, MDR2 maintains strong recognition performance while substantially reducing computational cost.
This setting reflects practical wearable applications, where reducing visual input is important for improving computational efficiency.

The main contributions of this paper include the following:
\begin{itemize}
    \item We propose MDR2, a teacher-guided sparse-to-dense multimodal pre-training framework for visual–inertial Ego-HAR. MDR2 supervises a shared fused representation with separate visual and inertial reconstruction targets, rather than a single joint target.

    \item We show that MDR2 consistently achieves higher mean F1-macro performance than the evaluated multimodal baselines and pre-training frameworks in the unified-backbone experiments on the EvIs-Kitchen dataset, providing empirical support for the effectiveness of dual unimodal reconstruction targets.

    \item We evaluate MDR2 under varying numbers of video key frames and show that it remains effective under sparse visual input, achieving strong performance in the four-key-frame setting with about 6.4\% of the computational cost of a spectrogram-based EviMAE encoder.
\end{itemize}

\vspace{-0.2cm}

\section{Related Work}
\vspace{-0.1cm}

\subsection{Multimodal Self-Supervised Learning}
Self-supervised learning (SSL) has emerged as a powerful alternative to supervised learning by alleviating the reliance on large-scale annotated data.
This advantage is particularly important in multimodal scenarios, where collecting ground-truth annotations is substantially more challenging than in unimodal settings due to cross-modal alignment and synchronization requirements \cite{bruce2022mmnet}, \cite{chen2015utd}, \cite{ma2022visual}, \cite{ma2023auto}.

Building upon unimodal successes (such as SimCLR~\cite{chen2020simCLR}, BYOL~\cite{grill2020byol}), recent studies have extended contrastive SSL frameworks to multimodal settings.
Radford \etal propose CLIP \cite{radford2021learning}, which aligns image and text representations via large-scale contrastive pre-training.
Owens \etal \cite{owens2018audio} exploit the temporal alignment between audio and video streams as a form of self-supervision.
Tsutsui \etal \cite{tsutsui2021you} also utilize the aligned segments of video clips and inertial sensor sequences as pairs to pre-train models.

Reconstruction-based multimodal SSL frameworks have also attracted increasing attention, as they avoid ambiguous pair construction in multimodal scenarios and offer more stable supervisory signals to guide representation learning.

Bao \etal \cite{bao2022vl} extend masked language modeling to jointly reconstruct image-text pairs, and subsequent works \cite{wang2023image} further enrich reconstruction targets to improve transferability among different downstream tasks.
Zhang \etal \cite{zhang2025evimae} propose EviMAE, which performs masked reconstruction pre-training on visual and inertial sensor data spectrograms.
To avoid tedious input-level reconstruction, Lian \etal \cite{lian2023av} extend Data2Vec to multimodal video-audio scenarios by performing reconstruction at the representation level.

Existing multimodal pre-training methods often rely on a single joint objective to learn shared representations across modalities.
Such designs provide a unified supervisory signal for different modalities, which may not fully accommodate the different representation-learning needs of modalities that capture different aspects of an activity \cite{dufumier2024align}, \cite{swetha2023presrving}.
This limitation is particularly evident in visual-inertial settings, where visual inputs primarily capture appearance and scene context, whereas inertial signals encode motion dynamics and patterns.

This observation motivates the exploration of multimodal pre-training objectives that do not collapse supervision into a single fused target.
In this work, we focus on this target-structure problem in reconstruction-based multimodal pre-training for visual-inertial Ego-HAR.

\vspace{-0.3cm}
\subsection{Cross-modal Learning}
Cross-modal learning aims to exploit complementary information across different modalities by transferring or distilling knowledge between them.

Several recent studies illustrate the effectiveness of cross-modal learning.
Qu \etal \cite{qu2022lipsound2} propose LipSound2, which leverages audio as supervision to guide lip video encoding through a reconstruction task, leading to improvements on video-audio tasks.
Tan \etal \cite{tan2023egodistill} distill knowledge from a heavyweight video encoder to a lightweight combination of an image encoder and an inertial sensor encoder.
Doždor \etal \cite{dovzdor2025sv} extend Data2Vec \cite{baevski2022data2vec} by incorporating skeleton data alongside video input during pre-training, which enables strong inference performance even without skeleton input.

Recently, several studies have explored leveraging unimodal signals as supervision to facilitate multimodal representation learning.
Du \etal \cite{du2023uni} propose guiding a multimodal encoder using well-trained unimodal encoders during joint supervised training.
Hsu \etal \cite{hsu2022uhubert} propose combining the unimodal and multimodal clustering tasks together to improve generalization ability.
Kim \etal \cite{kim2025cav2vec} introduce a unimodal auxiliary objective to enhance reconstruction-based pre-training.

Beyond multimodal recognition tasks, recent deep-learning studies in high-dimensional sensing, such as hyperspectral imaging \cite{hong2026hyperspectral}, have also explored modality heterogeneity and cross-modal fusion.
LRR-Net \cite{li2023lrr} uses interpretable deep unfolding for anomaly detection.
Im2State \cite{li2026im2state} models cross-state interactions for hyperspectral--multispectral fusion.
SpectralGPT \cite{hong2024spectralgpt} extends spatial--spectral representation learning through large-scale pretraining, while MambaX \cite{li2025mambax} introduces state predictive control for multimodal image super-resolution.
Although they address tasks different from visual-inertial Ego-HAR, they share the broader goal of explicitly modeling modality-specific structure during fusion. MDR2 follows this principle by using separate visual and inertial teacher targets.

The multimodal learning methods most closely related to MDR2 differ in their objectives and use of modality-specific guidance.
Cross-modal distillation methods typically transfer knowledge from one modality to another \cite{qu2022lipsound2}, whereas unimodal-teacher methods such as UMT \cite{du2023uni} use supervised modality-specific teachers to guide a multimodal model. 
Although we inherit the idea of using unimodal teachers from UMT, 
UMT uses this guidance to transfer modality-specific knowledge during downstream supervised multimodal training.
Its multimodal encoder is ultimately optimized with the label-prediction objectives and therefore requires labeled data throughout this supervised training process. 
In contrast, MDR2 is a representation pre-training framework that uses label-informed unimodal teachers to generate dense visual and inertial representation targets. 
The MDR2 reconstruction objective itself does not require activity labels for the pre-training samples, and could in principle be applied to additional unlabeled multimodal data beyond what is used in our current experiments.
Therefore, UMT is a label-dependent supervised downstream method, whereas MDR2 is a teacher-guided representation pre-training method with label-informed reconstruction targets.

\begin{figure*}[!t]
\centering
    \includegraphics[width=6.5in]{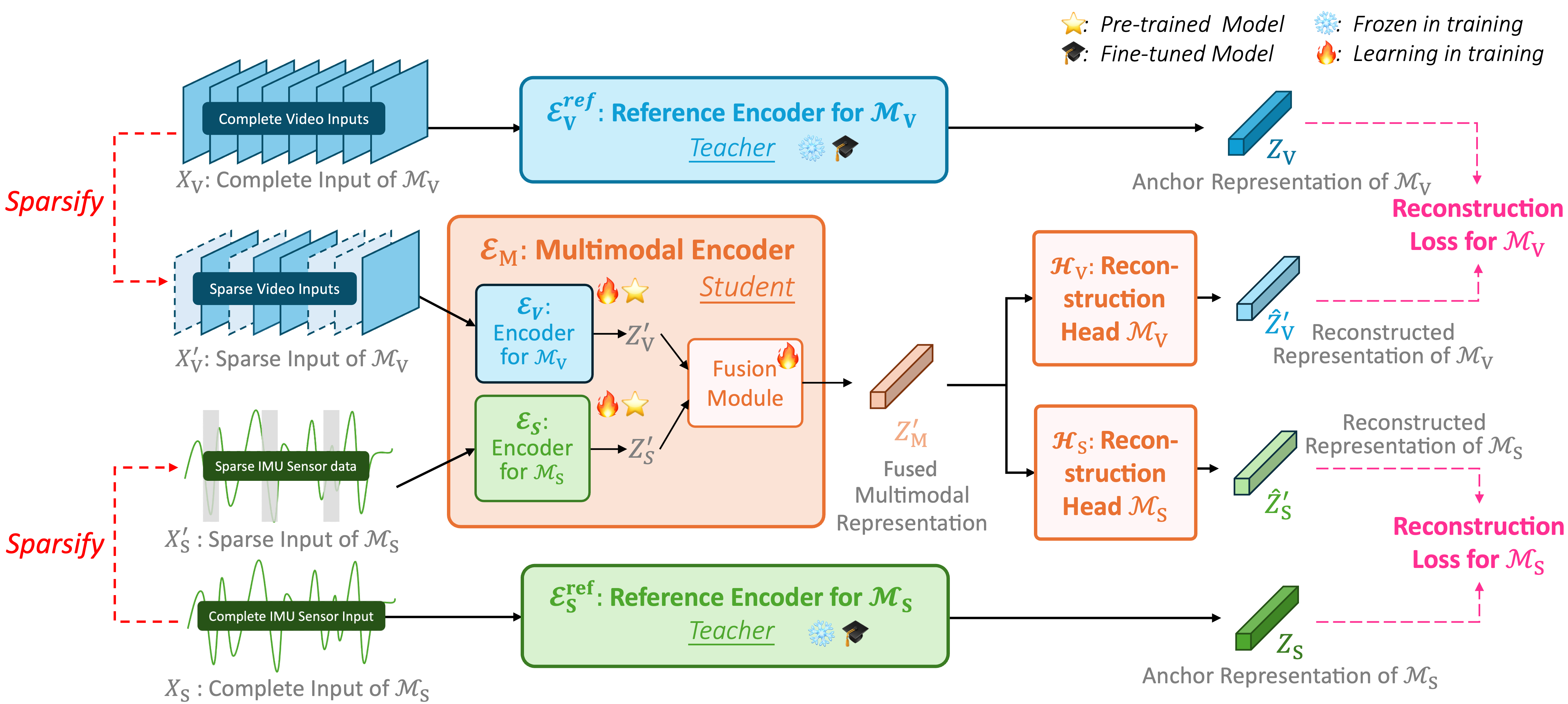}   
    \vspace{-0.1cm}
    \caption{Overview of the proposed teacher-guided multimodal representation pre-training framework, Multimodal Dual Representation Reconstruction (MDR2).
    A multimodal student encoder integrates sparse visual and inertial inputs into a fused representation.
    Two heads reconstruct the dense visual and inertial teacher representations from this fused representation.
    Instead of reconstructing toward a single shared target as in conventional representation reconstruction frameworks (Fig.~\ref{fig:ssl-rep-recon}), MDR2 supervises the fused representation using dual unimodal reconstruction targets.
    \label{fig:MDR2}
    }
\vspace{-0.3cm}
\end{figure*}

\subsection{Multi-target Learning}

Introducing multiple objectives or auxiliary supervision has been widely adopted to improve representation learning by providing complementary learning signals.

An example is multi-task learning (MTL) \cite{caruana1997multitask}, where multiple objectives jointly optimize shared representations. Prior studies have shown that auxiliary tasks can improve robustness and generalization across domains such as speech processing \cite{latif2022multitask} and video understanding \cite{baradaran2023multi}.
These findings suggest that multiple objectives can provide richer supervision than a single learning target.

Inspired by this observation, MDR2 adopts dual unimodal reconstruction objectives to provide separate visual and sensor-oriented supervisory signals during pre-training.

\section{Proposed Method}

\subsection{Overview}
To address the limitation of using a single joint supervision objective in multimodal pre-training, we propose \textbf{M}ultimodal \textbf{D}ual \textbf{R}epresentation \textbf{R}econstruction (MDR2), a teacher-guided pre-training framework that replaces joint supervision with dual unimodal reconstruction targets.

MDR2 formulates pre-training as a sparse-to-dense representation reconstruction task, as shown in Fig.~\ref{fig:MDR2}. 
The visual and inertial inputs are temporally sparsified to intentionally limit the information available in each modality and increase the reliance on cross-modal fusion during pre-training.
This sparse-input design is intended to improve recognition performance under reduced visual-input settings.

In parallel, we design a teacher-guided dual-target reconstruction mechanism based on the intuition that an effective fused representation should support reconstruction of unimodal reference representations.
Specifically, MDR2 uses two frozen unimodal teacher encoders to produce dense visual and inertial reference anchors.
The shared multimodal representation is then trained to reconstruct both anchors through separate visual and inertial reconstruction losses.
As a result, the multimodal encoder receives two modality-oriented supervisory signals, rather than a single shared target.

The remainder of this section introduces the detailed design of the MDR2 framework in Section~\ref{sec:method-MDR2}, followed by architectural details in Section~\ref{sec:method-arch}.

\subsection{Multimodal Dual  Representation Reconstruction}
\label{sec:method-MDR2}
We build our proposed framework upon two key design principles:
\begin{itemize}
    \item We design a sparse-to-dense reconstruction pre-training task that explicitly encourages the modeling of complementary information across modalities.
    \item We introduce a teacher-guided dual-target supervision mechanism, where visual and sensor reconstruction targets jointly supervise a shared multimodal encoder.
\end{itemize}
We use $\mathcal{M}_{\rm V}$ to represent the video (V) modality, and use $\mathcal{M}_{\rm S}$ to represent the IMU sensor (S) modality.

\subsubsection{Sparse-to-Dense Multimodal Encoding}
At this stage, we enable the two modalities to interact and model comprehensive and complementary information more effectively.

We artificially sparsify the inputs and feed them into the multimodal encoder $\boldsymbol{\mathcal{E}}_{\rm M}(\cdot)$ to extract the multimodal representation $Z'_{\rm M}$, which will be used consistently for the next unimodal representation reconstruction and downstream tasks.
\begin{equation}
    Z'_{\rm M} =  \boldsymbol{\mathcal{E}}_{\rm M}(
        X'_{\rm V}, X'_{\rm S}
    ),
\end{equation}

where $X'_{\rm V}$ and $X'_{\rm S}$ denote the sparse inputs of the V and S modalities, respectively.

\subsubsection{Teacher-Guided Dual-Target Reconstruction}
In this stage, we design a dual reconstruction-based supervision mechanism to guide the fused representation learned in the previous stage.

We formulate representation reconstruction from the sparse multimodal representation to the dense unimodal representations.
Two unimodal reconstruction heads $\boldsymbol{\mathcal{H}}_{\rm V}(\cdot)$ and $\boldsymbol{\mathcal{H}}_{\rm S}(\cdot)$ with identical architectures are applied to project the multimodal representation $Z'_{\rm M}$ into the corresponding unimodal representation spaces:
\begin{equation}
    \hat{Z}'_{\rm V} = \boldsymbol{\mathcal{H}}_{\rm V}( Z'_{\rm M} ), \quad
    \hat{Z}'_{\rm S} = \boldsymbol{\mathcal{H}}_{\rm S}( Z'_{\rm M} ),
\end{equation}

where $\hat{Z}'_{\rm V}$ and $\hat{Z}'_{\rm S}$ denote the reconstructed representations of each modality.
These projection heads map the fused representation into the corresponding visual and sensor reconstruction spaces and are discarded during downstream inference.

To supervise the reconstruction, we use two well-trained and frozen unimodal encoders $\boldsymbol{\mathcal{E}}^{\rm ref}_{\rm V}(\cdot)$ and $\boldsymbol{\mathcal{E}}^{\rm ref}_{\rm S}(\cdot)$ as reference models.
They take dense unimodal inputs and produce the reconstruction anchors:
\begin{equation}
    Z_{\rm V} = \boldsymbol{\mathcal{E}}^{\rm ref}_{\rm V}( X_{\rm V} ), \quad
    Z_{\rm S} = \boldsymbol{\mathcal{E}}^{\rm ref}_{\rm S}( X_{\rm S} ),
\end{equation}

Two unimodal reconstruction losses are defined as
\begin{equation}
    \mathcal{L}_{\rm V} = \mathcal{L}_{\rm MSE}(\hat{Z}'_{\rm V}, Z_{\rm V}), \quad
    \mathcal{L}_{\rm S} = \mathcal{L}_{\rm MSE}(\hat{Z}'_{\rm S}, Z_{\rm S}),
\end{equation}

and the total loss is the weighted sum of these unimodal losses:
\begin{equation}
    \mathcal{L}_{\rm total} = w_{\rm V} \cdot \mathcal{L}_{\rm V} + w_{\rm S} \cdot \mathcal{L}_{\rm S},
\end{equation}

The two reconstruction losses jointly supervise the fused multimodal representation during pre-training, with their relative contributions controlled by the loss weights $w_{\rm V}$ and $w_{\rm S}$.

We further analyze the effect of different loss weight settings in the ablation study (Section~\ref{subsec:abliation-weights}).

\subsection{Visual-Inertial Implementation Details}
\label{sec:method-arch}
We adopt the two-branch late-fusion architecture proposed in \cite{hao2024egocentric} as the multimodal encoder $\boldsymbol{\mathcal{E}}_{\mathrm{M}}(\cdot)$ for egocentric human activity recognition with visual input and dual-wrist inertial sensor data, denoted as V-S-S Ego-HAR.
It consists of a video encoder, a sensor encoder, and a fusion module, as the $\boldsymbol{\mathcal{E}}_{\rm M}(\cdot)$ shown in Fig.~\ref{fig:MDR2}.

\subsubsection{Pre-training Data Sparsification}
\label{subsec:sensor-sparsification}
During the pre-training, we apply temporal sparsification on the complete input data $X$ to obtain the sparse inputs $X'$. Both modalities are partially masked to encourage their cross-modal interaction and promote better multimodal complementarity modeling.
For the video data, we randomly select $N_{\rm p}$ frames from each action sample to reduce the temporal and motion information provided by the video stream.

\begin{figure}[!t]
    \centering{
    \includegraphics[width=3.3in]{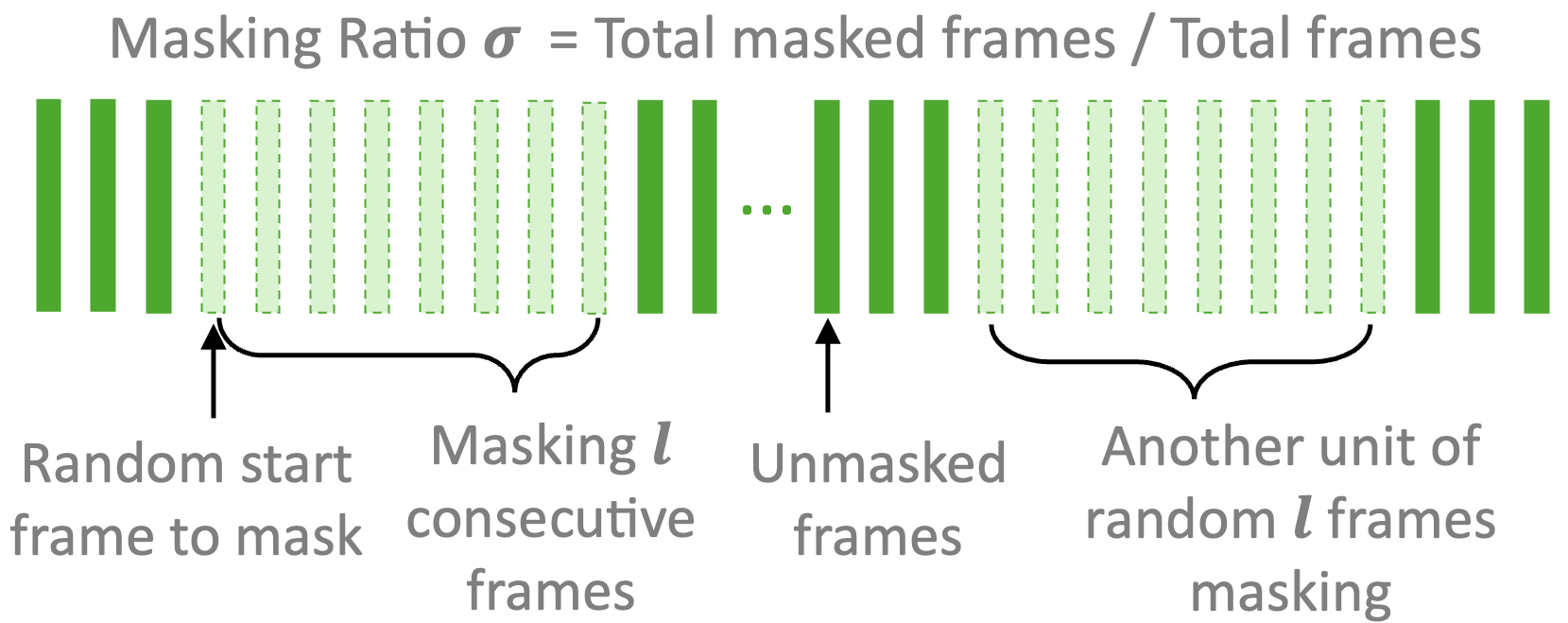}    
     \vspace{-0.2cm}
    \caption{Illustration of the sensor masking strategy, where each bar represents a single frame of inertial sensor data, and random masking is applied to clips of $l$ consecutive frames, with a total masking ratio $\sigma$.}
    \label{fig:sensor-mask-strategy}
    }
    \vspace{-0.3cm}
\end{figure}

For the inertial sensor data, we apply the clip-masking strategy. We mask the sequence with a unit of $l$ consecutive frames, and keep the total masking ratio as $\sigma$ (shown in Fig.~\ref{fig:sensor-mask-strategy}). 
Because a single frame of inertial sensor data contains limited semantic information, masking at the single-frame level would not introduce sufficient diversity between sparse and dense inputs.
The masking positions of the left and right sensors are identical. 
This design forces the multimodal encoder to rely on information from the other modality to compensate for missing segments, thereby improving the effectiveness of pre-training.

\subsubsection{Video Encoder}
We use Swin Video Transformer \cite{liu2022video} as our video encoder $\boldsymbol{\mathcal{E}}_{\rm V}$. With its 3-D shift-window strategy, it can extract both in-frame visual details and inter-frame visual motion, which provides an informative spatio-temporal video representation for further processing.
\begin{equation}
    Z_{\rm V} =  \boldsymbol{\mathcal{E}}_{\rm V}(X_{\rm V}),
\end{equation}

where $Z_{\rm V} \in \mathbb{R}^{D \times 1}$ denotes the video representation.

\subsubsection{Sensor Encoder}
We employ a transformer architecture to extract motion-related information from accelerometer and gyroscope data. 
Transformers are particularly effective at modeling long-range temporal dependencies in time-series data, and thus can benefit substantially from pre-training.

First, we concatenate all sensor data:
\begin{equation}
    X_{S} =  \text{Concat}(X_{\rm acc}, X_{\rm gyro}),
\end{equation}

where the dimensions of $X_{\rm acc}$ and $X_{\rm gyro}$ are $\mathbb{R}^{3 \times T \times 2}$, and the dimension of $X_{\rm S}$ is $\mathbb{R}^{6 \times T \times 2}$, where the last dimension denotes the left and right wrists, and $T$ denotes the total number of frames in the input sensor sequence.

Then we extract the sensor representation with our transformer encoder $\boldsymbol{\mathcal{E}}_{\rm S}$. It first embeds the raw sensor data into a high-dimensional space to capture correlations across sensor channels, then applies a self-attention mechanism \cite{vaswani2017attention} to encode the information.
\begin{equation}
    Z_{\rm S} =  \boldsymbol{\mathcal{E}}_{\rm S}(X_{\rm S}),
\end{equation}

where the final extracted sensor representation $Z_{\rm S}$ has the same shape as the video representation $Z_{\rm V}$.

\subsubsection{Fusion Module}
We apply a simple concatenation and linear projection layer to fuse two modalities.
\begin{equation}
    Z_{\rm M} =  \text{Linear}(\text{Concat}(Z_{\rm V}, Z_{\rm S})),
\end{equation}

To align with the unimodal representation, the final multimodal representation $Z_{\rm M}$'s dimension is also $\mathbb{R}^{D \times 1}$.
This representation is used in both pre-training and downstream supervised learning stages to maintain consistency between them.

\subsubsection{Reconstruction Head}
We use two three-layer MLPs to reconstruct the corresponding unimodal dense representations from the extracted multimodal representation:
\begin{equation}
    \hat{Z}_{\rm uni} = \boldsymbol{\mathcal{H}}_{\rm uni}(Z_{\rm M})  = \text{MLP}_{\rm uni}(Z_{\rm M}),
\end{equation}

where $\hat{Z}_{\rm uni} \in \mathbb{R}^{D \times 1}$, matching the dimension of $Z_{\rm M}$.

\section{Experiments}

\subsection{Experiment Settings}
\subsubsection{Dataset}
We select the EvIs-Kitchen dataset \cite{hao2023evis} as the primary dataset for training and evaluation because it provides well-synchronized ``V-S-S'' data, including egocentric videos and two-wrist sensor data, which directly matches the multimodal setting targeted by MDR2.
The two modalities exhibit distinct but complementary characteristics for activity understanding, and the synchronized recordings enable controlled evaluation with sparsely sampled video.
The dataset consists of cooking-related actions involving interactions between the subject and the environment.
Each action sample contains a verb label and a noun label, such as ``cut potato''.
The dataset contains 13.83 hours of data, of which 6.31 hours are annotated for verb- and noun-based activity recognition.

In addition, we collect additional two-wrist inertial sensor data, termed the ``\textbf{I}nertial \textbf{s}ensor daily \textbf{Act}ivity (IsAct)'' data extension.
It contains 19 hours of synchronized inertial sensor data from two wrists in the same format as the EvIs-Kitchen dataset. 
This extension is only used for sensor-only pre-training and provides additional unlabeled, domain-matched inertial data; it does not include annotations.

Following the subject-based protocol in \cite{hao2024egocentric}, we divide EvIs-Kitchen into four subject-disjoint folds.
Each fold contains three subjects and serves as the evaluation set in turn, while two of the remaining nine subjects are held out for validation and the others are used for training.
Hyperparameters are selected based on Fold 1 performance and then fixed across all experiments.
To avoid selection bias, only Folds 2--4 are included in the final performance aggregation.
For each evaluation fold, evaluation data are excluded from pre-training and teacher fine-tuning.
Thus, no evaluation samples or labels are used to construct the teacher targets.
Pre-training checkpoints are selected based on training loss, while downstream checkpoints are selected based on validation macro-F1.

Besides this main evaluation, we further use CMU-MMAC~\cite{de2009cmummac} for evaluating the compatibility of the proposed MDR2 objective when integrated into another video-IMU pre-training backbone.
We follow the evaluation setting used in~\cite{zhang2025evimae}.
The detailed results are described in Section~\ref{subsubsec:exp-compatibility}.

\subsubsection{Data Sparsification in Pre-training}
In our MDR2 pre-training, we select $N_{\rm p} = 4$ video frames per action sample, 
which retains basic visual observations while reducing the temporal information of the video input.
For inertial sensor data, we set the masking clip length $l = 8$ and the masking ratio $\sigma = 0.25$.
These operations increase the reconstruction difficulty and encourage the model to exploit complementary information across modalities.
During pre-training, the teacher encoders receive dense inputs: 32 video frames per action sample for the video teacher and unmasked inertial sensor data for the sensor teacher.
These settings were determined through a series of preliminary experiments.

Input sparsification with the above masking strategy is applied only during pre-training.
In downstream evaluation, we do not apply masking, while the number of input key frames $K$ is varied to evaluate performance under the evaluated sparse-input settings.

\subsubsection{Downstream Task and Metrics}
The downstream V-S-S Ego-HAR task consists of verb and noun recognition, which mainly reflect motion-related and visual understanding, respectively.
To evaluate recognition performance under sparse visual input, we train and evaluate the models with different numbers of video key frames, $K = 32, 16, 8, 4, 2$, where $K=32$ is treated as the full-observation setting.
The optimization settings follow \cite{hao2024egocentric}.
This evaluation examines whether pre-trained models maintain effective recognition performance under controlled reduced-frame training and evaluation settings, which are commonly encountered in realistic application scenarios.

For EvIs-Kitchen evaluation, we report F1-macro score, which is more suitable for datasets with imbalanced class distributions \cite{grandini2020metrics}.
Results are reported as fold-size-weighted means and descriptive standard deviations across the three evaluation folds (Folds 2--4), using the sum of weights as the variance denominator without sample-variance correction.
Paired comparisons use three equally weighted within-fold differences (MDR2 minus the comparison method), reported as mean $\pm$ sample SD, 95\% $t$-based confidence intervals, and unadjusted two-sided paired $t$-test $p$-values ($n=3$, $df=2$).
These analyses assume independent, approximately normally distributed fold differences.
Normality is difficult to assess with three folds. Therefore, the confidence intervals and tests are interpreted as approximate.

For CMU-MMAC evaluation, we follow the settings in \cite{zhang2025evimae} to report mean average precision (mAP).

\subsubsection{Training Procedure}
\label{subsubsec:training-procedure}
The pre-training procedure of MDR2 consists of three sequential stages:
\begin{enumerate}
    \item Unimodal initialization.
    \item Supervised fine-tuning of the teacher encoders.
    \item Multimodal pre-training with MDR2.
\end{enumerate}
After completing these stages, the multimodal student encoder is fine-tuned and evaluated on the downstream V-S-S Ego-HAR task in a supervised manner.
The unimodal teachers and reconstruction heads are used only during pre-training and are discarded during downstream evaluation, incurring no additional test-time overhead.

In the first stage, we initialize each unimodal encoder with modality-specific knowledge.
The video encoder directly adopts pre-trained weights from the Kinetics-400 dataset.
The sensor encoder is pre-trained from scratch on the IsAct data extension through the representation reconstruction task shown in Fig.~\ref{fig:ssl-rep-recon}.
Separate sensor teachers are trained from scratch for noun and verb recognition using EvIs-Kitchen training-split labels, with the same architecture as the sensor student encoder. 
After supervised training, each teacher is frozen to provide reconstruction targets during the corresponding sensor-only pre-training. The sensor teachers use the same supervised training settings as in Stage 2.
The masking strategy follows Section~\ref{subsec:sensor-sparsification}.
Sensor-only pre-training is performed for 100 epochs using AdamW with a batch size of 16, a base learning rate of $1 \times 10^{-3}$, and a cosine learning rate schedule with a 2.5-epoch warm-up starting at $5 \times 10^{-5}$.
This stage therefore uses teacher-guided pre-training rather than a fully label-free procedure.
Besides MDR2, all other methods in the unified-backbone comparison use the same video and sensor initialization settings.

Next, both unimodal encoders are initialized with the weights obtained in Stage 1 and fine-tuned using
EvIs-Kitchen activity labels to obtain the teacher models.
Teacher fine-tuning is performed separately for noun and verb recognition in each fold, using only the corresponding training-split labels.
The video teacher is fine-tuned for 50 epochs using AdamW with an initial learning rate of $1 \times 10^{-3}$ and a cosine learning rate schedule.
The sensor teacher is fine-tuned for 50 epochs using Adam with an initial learning rate of $2 \times 10^{-4}$, which is reduced by a factor of 10 every 15 epochs.
All experiments use a batch size of 16.

The fine-tuned encoders are then frozen and used as unimodal teacher reference models for the subsequent MDR2 pre-training stage.
To ensure a fair and controlled evaluation, the teacher encoders adopt the same backbone architectures as the corresponding student branches and the supervised multimodal baseline, so that the analysis focuses on the effect of teacher-guided dual-target reconstruction rather than differences in model capacity.
In this setting, MDR2 should be regarded as teacher-guided pre-training rather than fully label-free SSL: the reference teachers are fine-tuned using downstream activity labels from the training split of each fold, whereas the MDR2 reconstruction objective itself uses no activity labels.

In the third stage, MDR2 is used to pre-train the full multimodal encoder using EvIs-Kitchen training data without activity labels.
Both student branches are initialized with the weights obtained in Stage 1, while the teachers use the weights obtained in Stage 2.
The shared multimodal representation is supervised by two unimodal teacher targets, providing separate visual- and sensor-oriented reconstruction objectives.
MDR2 pre-training is performed for 100 epochs using AdamW with an initial learning rate of \(1 \times 10^{-3}\) and a cosine learning rate schedule with a batch size of 16.

\begin{table}[t!]
\centering
\setlength{\tabcolsep}{3.5pt}
\footnotesize
\renewcommand{\arraystretch}{1.2}

\caption{%
Downstream F1-macro performance (\%) under sparse ($K=4$)
and full ($K=32$) video observations. Results are reported as
weighted mean $\pm$ descriptive standard deviation across
three evaluation folds. Bold numbers indicate the best performance.}
\label{tab:mdr2-evis}

\scalebox{1.02}{%
\begin{tabular}{c|c|c|c|c}
\Xhline{3\arrayrulewidth}

\multirow{3}{*}{\textbf{Method}}
&
\multicolumn{4}{c}{\textbf{F1-Macro Score @1 (\%)}} \\

\cline{2-5}

&
\multicolumn{2}{c|}{\textbf{Noun Task}}
&
\multicolumn{2}{c}{\textbf{Verb Task}} \\

\cline{2-5}

&
\textbf{$K$=4}
&
\textbf{$K$=32}
&
\textbf{$K$=4}
&
\textbf{$K$=32} \\

\Xhline{3\arrayrulewidth}

\makecell*{V-Only \cite{liu2022video}\\ \scriptsize(no pretrain)}
& 63.99±\scriptsize0.74
& 80.15±\scriptsize1.37
& 61.40±\scriptsize0.72
& 75.21±\scriptsize0.79 \\

\hline

\makecell*{M-Baseline \cite{hao2024egocentric}\\ \scriptsize(no pretrain)}
& 65.89±\scriptsize0.48
& 80.27±\scriptsize1.21
& 66.47±\scriptsize0.88
& 76.05±\scriptsize1.49 \\

\hline

M-Contrastive \cite{owens2018audio}
& 64.18±\scriptsize1.36
& 79.39±\scriptsize0.98
& 68.13±\scriptsize1.43
& 75.12±\scriptsize1.09 \\

\hline

M-Data2Vec \cite{lian2023av}
& 66.18±\scriptsize1.03
& 77.20±\scriptsize0.99
& 67.92±\scriptsize1.34
& 76.12±\scriptsize2.05 \\

\hline

M-R2 \cite{hinton2015distilling}
& 69.56±\scriptsize2.11
& 79.96±\scriptsize1.39
& 70.36±\scriptsize0.83
& 75.29±\scriptsize1.25 \\

\hline

\textbf{MDR2 (proposed)}
& \textbf{75.09}±\scriptsize0.36
& \textbf{81.48}±\scriptsize0.67
& \textbf{72.64}±\scriptsize1.05
& \textbf{77.87}±\scriptsize1.61 \\[1pt]

\Xhline{3\arrayrulewidth}
\end{tabular}%
}

\vspace{-0.15cm}
\end{table}
\begin{table}[t!]
\centering
\setlength{\tabcolsep}{3pt}
\footnotesize
\renewcommand{\arraystretch}{1.3}

\caption{%
Paired differences in F1-macro performance between MDR2
and comparison methods.
Differences, SDs, and confidence intervals are expressed in
percentage points; positive differences favor MDR2.}
\label{tab:mdr2-paired}

\resizebox{\columnwidth}{!}{%
\begin{tabular}{c|c|c|c|c|c}
\Xhline{3\arrayrulewidth}

\multirow{2}{*}{\textbf{Comparison}}
&
\multirow{2}{*}{\textbf{Statistic}}
&
\multicolumn{2}{c|}{\textbf{Noun Task}}
&
\multicolumn{2}{c}{\textbf{Verb Task}} \\

\cline{3-6}

& &
\textbf{$K=4$}
&
\textbf{$K=32$}
&
\textbf{$K=4$}
&
\textbf{$K=32$} \\

\Xhline{3\arrayrulewidth}

\multirow{3}{*}{\makecell{MDR2 $-$\\V-Only}}
& \textbf{Mean diff. \scriptsize $\boldsymbol{\pm}$ SD}
& 11.10±\scriptsize1.18
& 1.34±\scriptsize1.63
& 11.24±\scriptsize2.14
& 2.67±\scriptsize1.85 \\

\cline{2-6}

& \textbf{95\% CI}
& $[8.16,\,14.04]$
& $[-2.72,\,5.40]$
& $[5.93,\,16.55]$
& $[-1.93,\,7.27]$ \\

\cline{2-6}

& \textbf{$\boldsymbol{p}$-value}
& 0.004
& 0.29
& 0.01
& 0.13 \\

\Xhline{3\arrayrulewidth}

\multirow{3}{*}{\makecell{MDR2 $-$\\M-Baseline}}
& \textbf{Mean diff. $\boldsymbol{\pm}$ SD}
& 9.20±\scriptsize0.91
& 1.22±\scriptsize1.23
& 6.17±\scriptsize2.01
& 1.83±\scriptsize1.35 \\

\cline{2-6}

& \textbf{95\% CI}
& $[6.94,\,11.46]$
& $[-1.84,\,4.27]$
& $[1.17,\,11.17]$
& $[-1.53,\,5.19]$ \\

\cline{2-6}

& \textbf{$\boldsymbol{p}$-value}
& 0.003
& 0.23
& 0.03
& 0.14 \\

\Xhline{3\arrayrulewidth}

\multirow{3}{*}{\makecell{MDR2 $-$\\M-R2}}
& \textbf{Mean diff. $\boldsymbol{\pm}$ SD}
& 5.52±\scriptsize2.42
& 1.53±\scriptsize1.09
& 2.28±\scriptsize2.30
& 2.58±\scriptsize2.85 \\

\cline{2-6}

& \textbf{95\% CI}
& $[-0.48,\,11.52]$
& $[-1.19,\,4.24]$
& $[-3.43,\,7.98]$
& $[-4.49,\,9.66]$ \\

\cline{2-6}

& \textbf{$\boldsymbol{p}$-value}
& 0.06
& 0.14
& 0.23
& 0.26 \\

\Xhline{3\arrayrulewidth}
\end{tabular}%
}
\vspace{-0.3cm}
\end{table}

\subsubsection{Baselines and Comparison Methods}
\label{sec:baselines}

We compare MDR2 with several supervised baselines and multimodal pre-training frameworks. 
All methods, including MDR2, are fine-tuned on the downstream V-S-S Ego-HAR task in a supervised manner for evaluation.
For supervised baselines, we include a video-only unimodal baseline (V-Only) and a multimodal baseline (M-Baseline).
The M-Baseline jointly trains the model using video and sensor inputs.
Both of them are trained and optimized in a fully supervised manner using labeled data on the downstream task.

For pre-training before downstream fine-tuning, we consider several representative multimodal frameworks covering different cross-modal learning paradigms:
\begin{itemize}
    \item M-Contrastive~\cite{owens2018audio} adopts a contrastive objective to align representations across modalities (illustrated in Fig.~\ref{fig:ssl-contra}). 
    It maximizes similarity between paired video and sensor samples and minimizes the similarity between unpaired samples.
    \item M-Data2Vec~\cite{lian2023av} follows an online teacher-student paradigm, in which the teacher parameters are updated as an exponential moving average (EMA) of the student parameters. 
    During training, the student takes masked data as inputs to predict the unmasked multimodal latent features generated by the teacher model.
    \item M-R2 follows a representation reconstruction paradigm (illustrated in Fig.~\ref{fig:ssl-rep-recon}) derived from knowledge distillation~\cite{hinton2015distilling}. 
    We obtain a multimodal teacher by supervised fine-tuning the joint-training M-Baseline on the downstream task.
    The teacher is then frozen to provide latent representations from complete multimodal inputs as supervision.
    The student model is trained to reconstruct these representations from masked multimodal inputs.
    This controlled comparison evaluates the combined change in teacher-target source and reconstruction structure under a shared student backbone and evaluation protocol.

\end{itemize}
In the unified-backbone comparison, all pre-training methods, including the proposed MDR2, share the same backbone architectures and downstream evaluation protocols.

We also include EviMAE~\cite{zhang2025evimae} as a recent video-IMU pre-training comparison method.
Unlike the above methods, EviMAE adopts spectrogram-based IMU inputs and different backbone structures.
Therefore, we report its results separately from the unified-backbone comparison, together with a computational metrics analysis.

To adapt EviMAE to EvIs-Kitchen, we modify its IMU input layer for the two wrist nodes with six channels per node ($x,y,z$ accelerometer and $x,y,z$ gyroscope).

For the ``EviMAE (Swin)'' variant in Tables~\ref{tab:evimae-evis-performance} and~\ref{tab:evimae-evis-efficiency}, we replace the original ViT video encoder with a Swin Video Transformer-Tiny.
No additional adaptation of the video input is required.
A separate linear layer projects the 768-dimensional Swin output features to 384 dimensions before they are passed to the original reconstruction components.

\begin{figure}[!t]
    \centering{
    \includegraphics[width=3.3in]{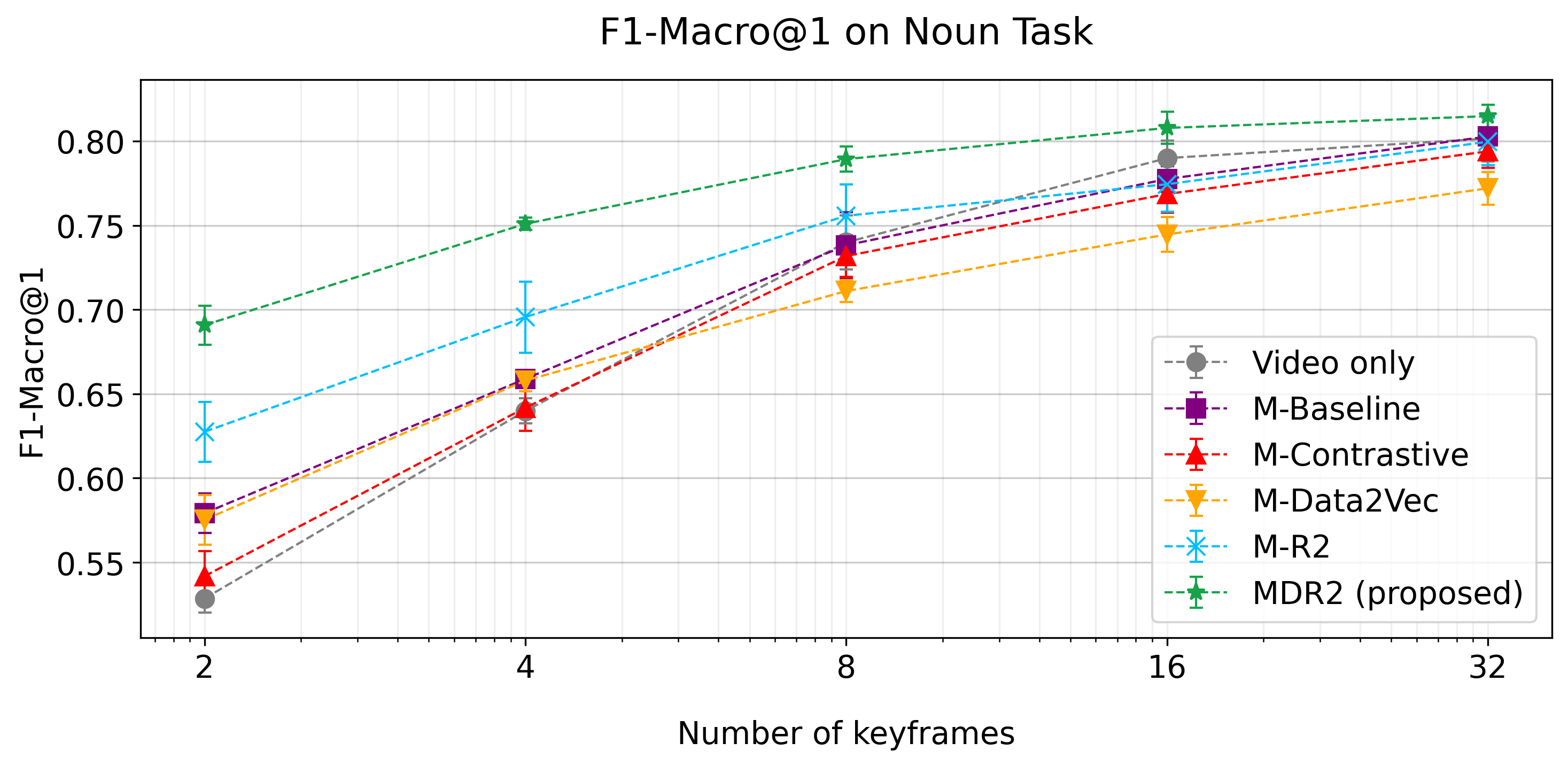}    
    \includegraphics[width=3.3in]{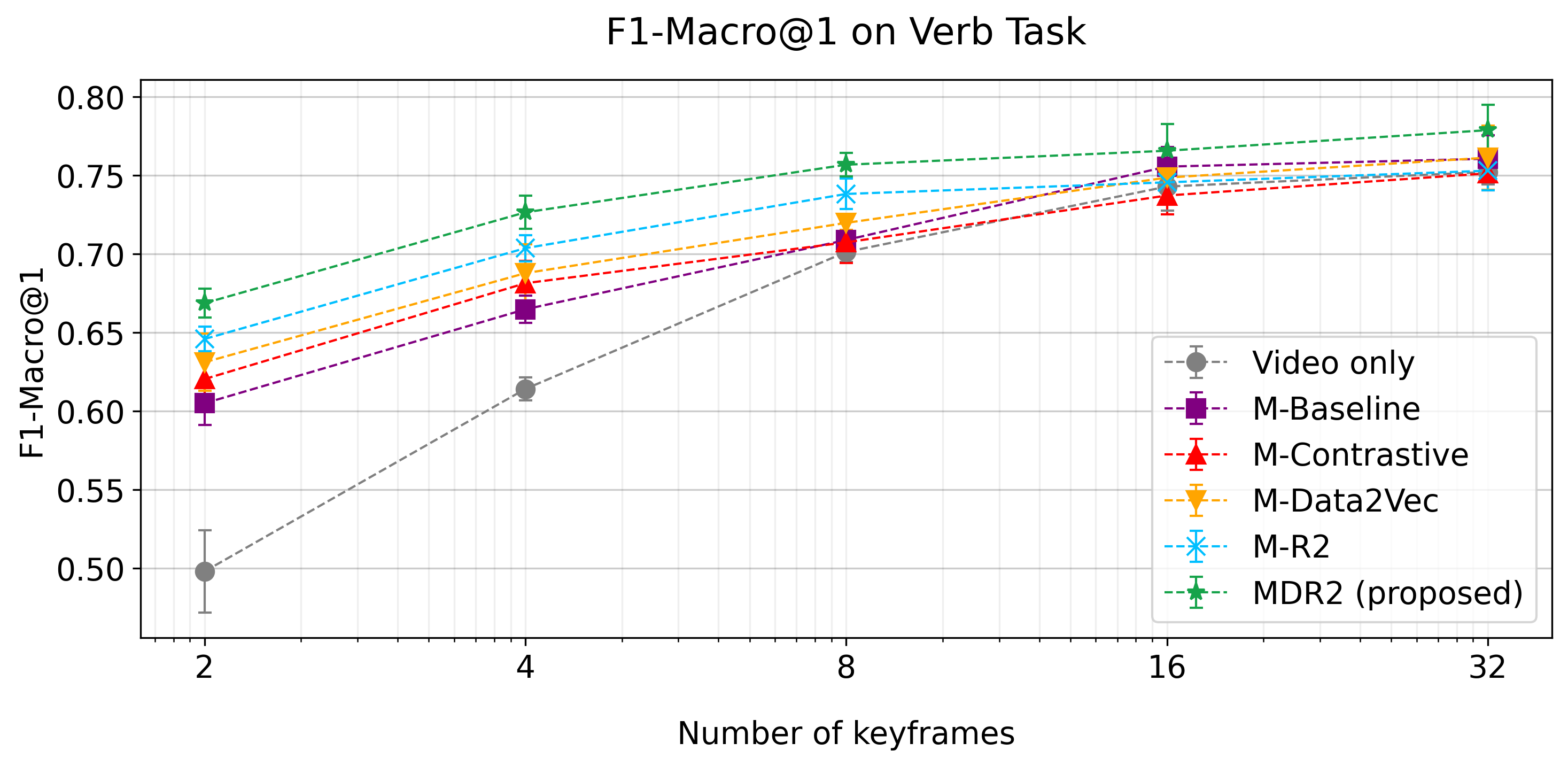}  
    \vspace{-0.3cm}
    \caption{Weighted mean F1-macro scores $\pm$ descriptive standard deviation across three evaluation folds (Folds 2--4) for different numbers of video key frames.}
    \label{fig:f1}
    }
    \vspace{-0.4cm}
\end{figure}

For the IMU branch, we retain the original device-based grouping, with each wrist treated as one device.
Each wrist provides six channels: three-axis acceleration and three-axis angular velocity. 
Following the original spectrogram preprocessing procedure, we generate six spectrograms per wrist instead of three and adapt the IMU input layer to accommodate the increased number of sensor patch tokens.
The IMU feature graph contains two nodes, corresponding to the left and right wrists. 
Initialization, preprocessing, and masking follow the original EviMAE settings~\cite{zhang2025evimae}, with one of the two wrist nodes randomly masked during graph masking.

For both the ViT- and Swin-based variants, all remaining pre-training and optimization settings follow the original EviMAE protocol~\cite{zhang2025evimae}.
Downstream evaluation follows the same three-fold protocol used for the other methods on EvIs-Kitchen.

\newcommand{\correct}[1]{%
  \textcolor{green!50!black}{\textbf{#1}}%
}

\newcommand{\incorrect}[1]{%
  \textcolor{red!75!black}{#1}%
}

\newcommand{\sampleblock}[5]{%
  \begin{minipage}[c]{3.2cm}
    \centering
    \includegraphics[width=0.76cm,height=0.76cm]{#1}\hspace{0.75pt}%
    \includegraphics[width=0.76cm,height=0.76cm]{#2}\hspace{0.75pt}%
    \includegraphics[width=0.76cm,height=0.76cm]{#3}\hspace{0.75pt}%
    \includegraphics[width=0.76cm,height=0.76cm]{#4}\\[-2pt]
    \scriptsize GT: #5
  \end{minipage}%
}

\begin{table*}[t]
\centering

\caption{Qualitative Top-3 prediction results under $K=4$ setting.
Percentages denote softmax probabilities.
Top-1 predictions are shown in green when correct and red when incorrect.
Correct lower-ranked predictions are shown in bold.}

\label{tab:qualitative}

\scriptsize
\setlength{\tabcolsep}{2pt}
\renewcommand{\arraystretch}{1.13}

\resizebox{\textwidth}{!}{%
\begin{tabular}{@{}p{3.2cm}cccccc@{}}

\toprule

\multirow[c]{2}{3.2cm}[-2pt]{%
  \centering
  \raisebox{6px}{\textbf{Sample Visualization}}%
}
& \multicolumn{2}{c}{\raisebox{3px}{\textbf{MDR2}}}
& \multicolumn{2}{c}{\raisebox{3px}{\textbf{M-R2}}}
& \multicolumn{2}{c}{\raisebox{3px}{\textbf{M-Baseline}}}
\\[-4pt]

\cmidrule(lr){2-3}
\cmidrule(lr){4-5}
\cmidrule(lr){6-7}

& \raisebox{2px}{\textbf{Verb Pred}} & \raisebox{2px}{\textbf{Noun Pred}}
& \raisebox{2px}{\textbf{Verb Pred}} & \raisebox{2px}{\textbf{Noun Pred}}
& \raisebox{2px}{\textbf{Verb Pred}} & \raisebox{2px}{\textbf{Noun Pred}}
\\[-3pt]

\midrule

\multirow{3}{*}{%
\sampleblock
{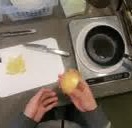}
{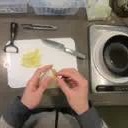}
{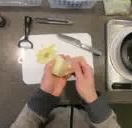}
{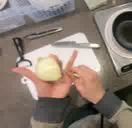}
{peel onion}}
& \correct{peel (93.90\%)}
& \correct{onion (53.31\%)}
& \incorrect{unpack (58.91\%)}
& \correct{onion (23.17\%)}
& \incorrect{unpack (92.17\%)}
& \incorrect{potato (82.55\%)}
\\

& unpack (6.10\%)
& garlic (38.11\%)
& \textbf{peel (40.43\%)}
& potato (22.68\%)
& \textbf{peel (7.70\%)}
& \textbf{onion (13.27\%)}
\\

& separate (\textless0.01\%)
& cheese (4.30\%)
& separate (0.56\%)
& banana (16.35\%)
& crack (0.05\%)
& garlic (3.84\%)
\\[3.5pt]

\multirow{3}{*}{%
\sampleblock
{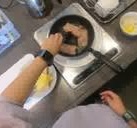}
{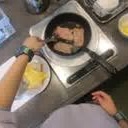}
{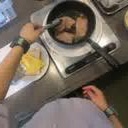}
{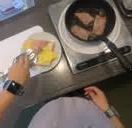}
{clamp bacon}}
& \correct{clamp (95.57\%)}
& \correct{bacon (99.70\%)}
& \correct{clamp (67.43\%)}
& \correct{bacon (78.17\%)}
& \incorrect{flip (61.24\%)}
& \correct{bacon (86.88\%)}
\\

& flip (3.52\%)
& egg (0.22\%)
& transfer (16.68\%)
& chicken (10.83\%)
& \textbf{clamp (24.12\%)}
& chicken (2.41\%)
\\

& stir (0.75\%)
& steak (0.03\%)
& flip (11.35\%)
& egg (0.65\%)
& squeeze (7.12\%)
& cheese (0.57\%)
\\[3.5pt]

\multirow{3}{*}{%
\sampleblock
{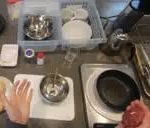}
{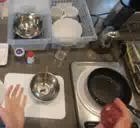}
{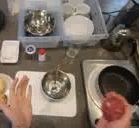}
{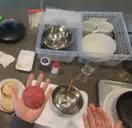}
{throw-around patty}}
& \correct{throw-around (99.95\%)}
& \correct{patty (99.64\%)}
& \correct{throw-around (93.49\%)}
& \incorrect{hand (53.12\%)}
& \correct{throw-around (92.14\%)}
& \incorrect{banana (42.72\%)}
\\

& wipe (0.02\%)
& hand (0.27\%)
& peel (4.40\%)
& \textbf{patty (26.86\%)}
& squeeze (6.11\%)
& strawberry-jam (40.98\%)
\\

& shape (0.01\%)
& steak (0.02\%)
& wipe (1.79\%)
& chicken (3.29\%)
& sprinkle (0.92\%)
& potato (5.97\%)
\\[2pt]

\bottomrule

\end{tabular}%
}
\vspace{-0.5cm}
\end{table*}

\vspace{-0.15cm}
\subsection{Comparisons with Multimodal Pre-training Methods}
\subsubsection{Comparison under the Unified Backbone Setting}
Table~\ref{tab:mdr2-evis} summarizes the downstream Ego-HAR evaluation results under the sparse-input setting ($K=4$) and the full-observation setting ($K=32$).
Fig.~\ref{fig:f1} further shows the F1-macro trends across different values of $K$.
Table~\ref{tab:mdr2-paired} complements these results with paired performance differences, 95\% confidence intervals, and $p$-values for the main comparisons.

In the unified-backbone comparison, MDR2 achieves the highest mean F1-macro scores among the methods in Table~\ref{tab:mdr2-evis} for both tasks at $K=4$ and $K=32$.
Its gains over M-Baseline and V-Only are larger under the sparse-input setting ($K=4$) than under $K=32$.
The comparison with M-R2 evaluates the combined effect of MDR2’s teacher-target selection and reconstruction structure against single-target multimodal reconstruction.
Under the same unified backbone and evaluation protocol, MDR2 consistently achieves higher mean F1-macro scores than M-R2. 
In particular, when $K$=4, MDR2 improves the F1-macro score from 69.56\% to 75.09\% on the noun task and from 70.36\% to 72.64\% on the verb task,
and Table~\ref{tab:mdr2-paired} also reports the corresponding unweighted mean paired improvements of 5.52 percentage points on the noun task (95\% CI: [-0.48, 11.52]) and 2.28 percentage points on the verb task (95\% CI: [-3.43, 7.98]).
Although both comparisons show positive mean improvements, neither reaches statistical significance at the 0.05 level.

Table~\ref{tab:qualitative} presents qualitative top-3 predictions for three activity samples at K=4. In these examples, MDR2 predicts both the verb and noun correctly at rank 1, whereas M-R2 and M-Baseline make errors in some cases.

These observed mean improvements suggest benefits from MDR2's combined teacher-target and reconstruction design under sparse-input conditions.
To further examine reconstruction-head parameterization under fixed teacher targets, Table~\ref{tab:joint-target} presents a controlled comparison using the same two unimodal teachers.

These performance trends are consistent with two key design choices of MDR2.
First, the unimodal teacher encoders provide two modality-oriented supervision signals, allowing the shared multimodal representation to be optimized toward both visual and sensor reconstruction targets.
Second, the sparse-to-dense reconstruction setting increases the difficulty of the pre-training task, encouraging the model to exploit complementary cues across modalities when reconstructing dense teacher representations from sparse multimodal inputs.

Overall, MDR2 achieves higher mean downstream performance than the unified-backbone baselines on both noun and verb recognition across the evaluated key-frame settings, with larger relative gains in the lower-frame settings.

\begin{table}[!t]
\centering
\caption{Comparison with EviMAE \cite{zhang2025evimae} on downstream performance.
Bold and underlined numbers indicate the best and second-best results.}
\label{tab:evimae-evis-performance}
\renewcommand{\arraystretch}{1.3}
\scalebox{1}{
\begin{tabular}{c|c|cc|cc}
\Xhline{3\arrayrulewidth}

\multirow{3}{*}{\makecell{\textbf{Model}\\\textbf{Structure}}}
&
\multirow{3}{*}{\makecell{\textbf{Pretrain}\\\textbf{Method}}}
&
\multicolumn{4}{c}{\textbf{F1-Macro Score @1 (\%)}} \\

\cline{3-6}

&
&
\multicolumn{2}{c|}{\textbf{Noun Task}}
&
\multicolumn{2}{c}{\textbf{Verb Task}}
\\

\cline{3-6}

&
&
$\mathbf{K=4}$
&
$\mathbf{K=32}$
&
$\mathbf{K=4}$
&
$\mathbf{K=32}$
\\

\Xhline{3\arrayrulewidth}

\multirow{2}{*}{\makecell{EviMAE\\(ViT) \cite{zhang2025evimae}}
}
&
-
&
44.35
&
57.88
&
55.07
&
63.59
\\

\cline{2-6}

&
EviMAE
&
49.20
&
61.23
&
57.18
&
65.16
\\

\hline

\multirow{2}{*}{\makecell{EviMAE\\(Swin)}
}
&
-
&
65.49
&
80.55
&
67.81
&
77.53
\\

\cline{2-6}

&
EviMAE
&
\underline{67.08}
&
\textbf{81.84}
&
\underline{68.99}
&
\textbf{78.26}
\\

\hline

\multirow{2}{*}{Ours}
&
-
&
65.89
&
80.27
&
66.47
&
76.05
\\

\cline{2-6}

&
MDR2
&
\textbf{75.09}
&
\underline{81.48}
&
\textbf{72.64}
&
\underline{77.87}
\\

\Xhline{3\arrayrulewidth}

\end{tabular}
}
\vspace{-0.3cm}
\end{table}

\subsubsection{Comparison with Different Backbone Structures}
\label{subsubsec:different-backbone-comparison}

Table~\ref{tab:evimae-evis-performance} compares MDR2 with EviMAE in terms of downstream recognition performance under different visual input settings. EviMAE uses spectrogram-based IMU inputs and different visual backbones, including ViT and Swin.

Table~\ref{tab:evimae-evis-efficiency} summarizes the model size, computational cost, and inference latency of MDR2 and EviMAE.
Computational cost and inference latency were measured on a single NVIDIA A100 GPU with a batch size of 16. 
Latency is reported as the average per-sample forward-pass time, excluding data transfer.
GMACs and inference latency were measured for both $K=4$ and $K=32$ using the same settings for all models.

Under the full-observation setting ($K=32$), MDR2 achieves performance comparable to EviMAE-Swin while using only 30.5M parameters, compared with 111.3M for EviMAE-Swin.
Moreover, MDR2 requires substantially lower computational cost (18.7 vs. 58.1 GMACs) and lower inference latency (4.19 vs. 7.45 ms/sample), demonstrating a more efficient inference process.

The advantage becomes more evident under the sparse-input setting ($K=4$).
MDR2 outperforms both EviMAE variants on noun and verb recognition while reducing the computational cost to only 2.6 GMACs, approximately 6.4\% of that of EviMAE-Swin.
This reduction in computational cost also translates into the lowest inference latency (1.07 ms/sample), indicating that MDR2 achieves a favorable trade-off between recognition performance and computational efficiency, especially when only a small number of visual key frames are available.

\subsubsection{Compatibility with Another Backbone and Dataset}
\label{subsubsec:exp-compatibility}
To further examine the compatibility of the proposed objective, we integrate MDR2 into the EviMAE framework and evaluate it on CMU-MMAC.
In this experiment, the EviMAE backbone and input preprocessing are kept unchanged. The original EviMAE losses are retained, and the MDR2 reconstruction loss, with equal weights for the two modalities, is added to the pre-training objective without an additional scaling factor.
The two unimodal teachers use the same video and IMU encoder architectures as EviMAE and are fine-tuned using CMU-MMAC activity labels following EviMAE's unimodal supervised training protocol~\cite{zhang2025evimae}.
The teacher representation targets supervise the multimodal representation produced by the unified encoder before the decoder stage. This multimodal representation is also used for downstream evaluation.

As shown in Table~\ref{tab:evimae-cmummac}, adding the MDR2 objective to the EviMAE backbone further improves the mAP from the reproduced baseline score of 87.76\% to 88.92\%. 
This improvement reflects the combined effect of activity-label-based teacher adaptation and dual-target
reconstruction within the EviMAE pre-training framework.
Overall, this compatibility experiment suggests that the proposed objective can be integrated into another video-IMU pre-training backbone and provide complementary performance gains.
We regard this result as preliminary evidence of portability, while broader validation across more diverse activity domains remains future work.

\begin{table}[!t]
\centering
\caption{Comparison with EviMAE \cite{zhang2025evimae} in model size, computational cost, and inference latency.
Lower values indicate higher computational efficiency.}
\label{tab:evimae-evis-efficiency}
\renewcommand{\arraystretch}{1.3}
\scalebox{0.88}{
\begin{tabular}{c|c|cc|cc}
\Xhline{3\arrayrulewidth}

\multirow{2}{*}{\makecell{\textbf{Model}\\\textbf{Structure}}}
&
\multirow{2}{*}{\makecell{\textbf{Model}\\\textbf{Size} (M)}}
&
\multicolumn{2}{c|}{\makecell{\rule{0pt}{2.5ex}\textbf{Computational Cost}\\(GMACs)}}
&
\multicolumn{2}{c}{\makecell{\rule{0pt}{2.5ex}\textbf{Inference Latency}\\(ms/sample)}}
\\

\cline{3-6}

&
&
$\mathbf{K=4}$
&
$\mathbf{K=32}$
&
$\mathbf{K=4}$
&
$\mathbf{K=32}$
\\

\Xhline{3\arrayrulewidth}

EviMAE (ViT) \cite{zhang2025evimae}
&
104.1
&
39.9
&
52.4
&
2.82
&
6.27
\\

\hline

EviMAE (Swin)
&
111.3
&
40.6
&
58.1
&
3.14
&
7.45
\\

\hline

Ours
&
\textbf{30.5}
&
\textbf{2.6}
&
\textbf{18.7}
&
\textbf{1.07}
&
\textbf{4.19}
\\

\Xhline{3\arrayrulewidth}

\end{tabular}
}
\vspace{-0.3cm}
\end{table}
\begin{table}[!t]
\centering
\caption{Compatibility of MDR2 with the EviMAE framework on the CMU-MMAC dataset.
Bold numbers indicate the best performance.}
\label{tab:evimae-cmummac}
\renewcommand{\arraystretch}{1.25}
\setlength{\tabcolsep}{5pt}

\begin{tabular}{c|c|c}
\Xhline{3\arrayrulewidth}

\textbf{Model Structure}
&
\textbf{Pretrain Framework}
&
\textbf{mAP (\%)}
\\

\Xhline{3\arrayrulewidth}

\multirow{3}{*}{\makecell{EviMAE (ViT)\\\textit{Original Structure}}}

&
-
&
73.28
\\

\cline{2-3}

&
EviMAE
&
87.76
\\

\cline{2-3}

&
EviMAE + MDR2
&
\textbf{88.92}
\\

\Xhline{3\arrayrulewidth}

\end{tabular}

\vspace{-0.3cm}
\end{table}

\subsection{Ablation Experiments on MDR2}

\subsubsection{Comparison with Joint-target Head Design}

To examine the effect of the reconstruction head design within the teacher-guided framework, we compare MDR2 with a joint-target reconstruction variant using the same two fine-tuned unimodal teachers, student encoder, and training protocol.
Both teacher targets are 768-dimensional and used without additional normalization. Both variants employ element-wise mean MSE with equal weighting across target elements, with $w_V=w_S=0.5$ for MDR2.
The reconstruction MLPs use ReLU activations in their hidden layers.

The joint variant predicts the concatenated 1536-dimensional target using a
$768\!\!\rightarrow\!\!768\!\!\rightarrow\!\!1536\!\!\rightarrow\!\!1536$
MLP with 4.13M parameters, whereas MDR2 uses two
$768\!\!\rightarrow\!\!768\!\!\rightarrow\!\!768\!\!\rightarrow\!\!768$
modality-specific MLPs with 3.54M parameters in total.

For identical predictions and targets, the mean MSE on the concatenated target equals the average of the two modality-specific mean MSE losses. Splitting the loss into two terms therefore does not itself introduce additional supervisory information.
Thus, the comparison primarily concerns shared versus modality-specific reconstruction-head parameterization.

The results are shown in Table~\ref{tab:joint-target}.

Compared with the joint-head variant, 
MDR2 achieves higher mean downstream performance under both sparse-input ($K=4$) and full-observation ($K=32$) settings.
The mean paired improvements range from 0.26 to 1.12 percentage points, but 95\% confidence intervals in all settings include zero and all $p$-values exceed 0.05.
These results indicate a favorable mean performance trend for the modality-specific-head design, but do not establish statistically significant superiority.

One possible explanation is that the joint head shares its nonlinear mapping across the concatenated visual and sensor output spaces, whereas MDR2 uses independently parameterized heads for the two representation spaces.
This separation may allow each head to adapt its mapping to the corresponding teacher representation while both reconstruction objectives optimize the shared multimodal representation.

\renewcommand{\arraystretch}{1.3}

\begin{table}[t!]
\centering
\caption{
Comparison of joint-head and dual-target reconstruction
using the same teachers and training protocol.
Performance and paired statistics follow
Tables~\ref{tab:mdr2-evis} and~\ref{tab:mdr2-paired}, respectively.
Paired differences are MDR2 minus joint-head.
Bold indicates the best performance.}
\vspace{-0.15cm}
\label{tab:joint-target}

\scalebox{0.85}{%
\begin{tabular}{c|c|c@{\hspace{8pt}}c|c@{\hspace{8pt}}c}
\Xhline{3\arrayrulewidth}

\multicolumn{2}{l|}{%
\multirow{2}{*}{\textbf{Target Formulation}}}
&
\multicolumn{2}{c|}{\textbf{Noun Task}}
&
\multicolumn{2}{c}{\textbf{Verb Task}} \\

\cline{3-6}

\multicolumn{2}{c|}{}
&
$K=4$
&
$K=32$
&
$K=4$
&
$K=32$ \\

\Xhline{3\arrayrulewidth}

\multicolumn{2}{l|}{Joint-concatenated target}
& 73.97±\scriptsize0.52
& 81.23±\scriptsize0.53
& 72.01±\scriptsize0.60
& 77.27±\scriptsize1.19 \\

\multicolumn{2}{l|}{Dual targets (MDR2)}
& \textbf{75.09}±\scriptsize0.36
& \textbf{81.48}±\scriptsize0.67
& \textbf{72.64}±\scriptsize1.05
& \textbf{77.87}±\scriptsize1.61 \\

\Xhline{3\arrayrulewidth}

\multirow{3}{*}{\makecell{\textbf{Paired}\\\textbf{Diff.}}}
& \textbf{Mean $\boldsymbol{\pm}$ SD}
& 1.12±\scriptsize0.99
& 0.26±\scriptsize0.59
& 0.63±\scriptsize2.02
& 0.61±\scriptsize0.51 \\

\cline{2-6}

& \textbf{95\% CI}
& [-1.35,\,3.59]
& [-1.20,\,1.71]
& [-4.39,\,5.64]
& [-0.67,\,1.88] \\

\cline{2-6}

& \textbf{$\boldsymbol{p}$-value}
& 0.19
& 0.53
& 0.64
& 0.18 \\

\Xhline{3\arrayrulewidth}
\end{tabular}%
}

\vspace{-0.2cm}
\end{table}

\subsubsection{Learning Policy and Teacher Quality Comparison}
We investigate the effects of learning policy and teacher quality in multimodal pre-training:
\begin{enumerate}
    \item[i] \textbf{Student Update Strategy}: This refers to whether we update the unimodal branches $\boldsymbol{\mathcal{E}_{\rm uni}}$ of the student model $\boldsymbol{\mathcal{E}_{\rm M}}$, or freeze them and only update the fusion module.
    \item[ii] \textbf{Teacher Quality}: This refers to using either a teacher model  with unimodal pre-training only (Step 1 in Section \ref{subsubsec:training-procedure}) or a teacher model further fine-tuned through supervised learning (Step 2 in Section \ref{subsubsec:training-procedure}).
\end{enumerate}
The results are shown in Table~\ref{tab:leaning-policy}.

We first consider the baseline setting (Exp~① in Table~\ref{tab:leaning-policy}), where the unimodal encoders in the student model are frozen and supervision is provided by pre-trained unimodal teachers.
Relative to M-Baseline in Table~\ref{tab:mdr2-evis}, this configuration yields mixed results, with higher mean scores for nouns at K=4 and verbs at K=32, but lower mean scores in the other two settings. Freezing the student branches and using Stage 1 teacher representations therefore does not provide consistent gains across the evaluated settings.

\renewcommand{\arraystretch}{1.2} 
\begin{table}[t!]
\centering
\caption{Learning Policy and Teacher Quality Comparisons of MDR2.
Bold numbers indicate the best performance.}
\vspace{-0.17cm}
\scalebox{0.83}{
\label{tab:leaning-policy}
\begin{tabular}{>{\centering\arraybackslash}m{0.4cm}|>{\centering\arraybackslash}m{1.8cm}>{\centering\arraybackslash}m{2cm}|rrrr}
\Xhline{3\arrayrulewidth}
\multirow{3}{*}{\centering
\makecell{\textbf{Exp}\\\textbf{ID}}
} 
& \multicolumn{2}{c|}{\textbf{Learning Policy}} 
& \multicolumn{4}{c}{\textbf{F1-Macro Score @1 (\%)}} \\ 
\cline{2-7} 
& \multicolumn{1}{c|}{\multirow{2}{*}{\centering
\makecell{\textbf{Update $\boldsymbol{\mathcal{E}_{\rm uni}}$ in}\\ \textbf{Student Model?}}
}} 
& \multirow{2}{*}{\centering
\makecell{\textbf{Use Fine-tuned}\\ \textbf{Teacher Model?}}
} 
& \multicolumn{2}{c|}{\textbf{Noun Task}} 
& \multicolumn{2}{c}{\textbf{Verb Task}} \\ 
\cline{4-7} 
& \multicolumn{1}{c|}{} 
&  
& $K$=$4$
& \multicolumn{1}{c|}{$K$=$32$}
& $K$=$4$
& $K$=$32$ \\ 
\Xhline{3\arrayrulewidth}

① & \multicolumn{1}{c|}{$\times$} & $\times$
& 66.43 & \multicolumn{1}{c|}{79.42} & 65.57 & 76.19 \\

② & \multicolumn{1}{c|}{$\checkmark$} & $\times$
& 67.62 & \multicolumn{1}{c|}{76.68} & 69.18 & 75.37 \\

③ & \multicolumn{1}{c|}{$\times$} & $\checkmark$
& 66.19 & \multicolumn{1}{c|}{79.23} & 66.65 & 75.70 \\

④ & \multicolumn{1}{c|}{$\checkmark$} & $\checkmark$
& \textbf{75.09} & \multicolumn{1}{c|}{\textbf{81.48}}
& \textbf{72.64} & \textbf{77.87} \\

\Xhline{3\arrayrulewidth}
\end{tabular}
}
\vspace{-0.4cm}
\end{table}

When the unimodal encoders in the student model are allowed to update without fine-tuned teachers (Exp~② in Table~\ref{tab:leaning-policy}), performance improves for both tasks under the $K=4$ setting but decreases for both tasks under the $K=32$ setting.
Thus, updating the student unimodal encoders alone does not yield consistent gains across the evaluated visual-observation settings.

Importantly, introducing a fine-tuned teacher while keeping the student unimodal encoders frozen (Exp~③ in Table~\ref{tab:leaning-policy}) does not yield clear performance gains.
This result suggests that improved teacher representations alone are insufficient, and that effective knowledge transfer in MDR2 benefits from allowing the student encoders to adapt to the reconstruction objectives.

Together, these results indicate that combining fine-tuned teachers with updates to the student unimodal encoders (Exp~④ in Table~\ref{tab:leaning-policy}) yields the best performance among the evaluated MDR2 configurations.

\renewcommand{\arraystretch}{1.2} 
\begin{table}[t!]
\centering
\caption{Input Modality Sparsification Comparison on MDR2.
Bold numbers indicate the best performance.}
\vspace{-0.17cm}
\label{tab:mod-mask}
\begin{tabular}{c|c|cccc}
\Xhline{3\arrayrulewidth}
\multirow{3}{*}{\makecell{\textbf{Exp} \\ \textbf{ID}}} & \multirow{3}{*}{\makecell{\textbf{Sparsified}\\ \textbf{Input Modality}}} 
& \multicolumn{4}{c}{\textbf{F1-Macro Score @1 (\%)}} \\ \cline{3-6} 
& & \multicolumn{2}{c}{\textbf{Noun Task}} & \multicolumn{2}{c}{\textbf{Verb Task}} \\ \cline{3-6} 
& & $K=4$ & \multicolumn{1}{c|}{$K=32$} & $K=4$ & $K=32$ \\
\Xhline{3\arrayrulewidth}
① & None
& 70.63 & \multicolumn{1}{c|}{79.81} & 68.73 & 76.28 \\

② & Video-only
& 74.66 & \multicolumn{1}{c|}{80.07} & 71.83 & 77.13 \\

③ & Sensor-only
& 71.08 & \multicolumn{1}{c|}{80.12} & 69.09 & 76.63 \\

④ & Both
& \textbf{75.09} & \multicolumn{1}{c|}{\textbf{81.48}}
& \textbf{72.64} & \textbf{77.87} \\

\Xhline{3\arrayrulewidth}
\end{tabular}
\vspace{-0.5cm}
\end{table}

\subsubsection{Input Modality Sparsification Comparison}

We investigate the effect of applying input sparsification to different modalities in the MDR2 framework.
The results are shown in Table~\ref{tab:mod-mask}.

Without input sparsification (Exp~① in Table~\ref{tab:mod-mask}), the model obtains the lowest average performance among the evaluated MDR2 sparsification variants.
When both modalities are provided with dense inputs, the reconstruction objective may rely more on within-modality cues, reducing the need for cross-modal interaction during pre-training.

When sparsification is applied to only one modality (Exp~② and ③ in Table~\ref{tab:mod-mask}), the performance shows only limited improvement.
In this case, the sparsified modality benefits from cross-modal information, while the dense modality continues to provide sufficient cues for reconstruction.

These results indicate that sparsifying only one modality provides smaller gains than sparsifying both modalities within the MDR2 framework.
Compared with sensor-only sparsification, video-only sparsification generally yields larger improvements, particularly under the sparse visual-input setting, where reduced visual input increases reliance on multimodal cues.

When both modalities are sparsified (Exp~④ in Table~\ref{tab:mod-mask}), the model achieves the best overall performance.
Under this setting, neither modality provides complete temporal observations, making cross-modal cues more useful for reconstructing dense teacher representations.
Consequently, the model is encouraged to use complementary cues from both modalities when optimizing the dual reconstruction objectives.

Within the MDR2 framework, sparsifying both modalities yields the best overall performance among the evaluated sparsification strategies and may encourage multimodal interaction during pre-training, improving downstream performance under reduced visual-input settings.

\begin{table}[t!]
\centering
\caption{Input Temporal Sparsity Comparisons on MDR2. \\
Bold numbers indicate the best performance.}
\vspace{-0.17cm}
\label{tab:SSL-inputs}
\begin{tabular}{c|cc|cccc}
\Xhline{3\arrayrulewidth}
    \multirow{3}{*}{\makecell{\textbf{Exp} \\ \textbf{ID}}}
    & \multicolumn{2}{c|}{\textbf{Input Sparsity}} 
    & \multicolumn{4}{c}{\textbf{F1-Macro Score @1 (\%)}} \\
\cline{2-7} 

    & \multirow{2}{*}{\makecell{\textbf{Video} \\ ($N_{\rm p}$)}} 
    & \multirow{2}{*}{\makecell{\textbf{Sensor} \\ ($\sigma$)}} 
    & \multicolumn{2}{c|}{\textbf{Noun Task}} 
    & \multicolumn{2}{c}{\textbf{Verb Task}} \\
\cline{4-7} 
    & & & \textbf{$K=4$} & \multicolumn{1}{c|}{\textbf{$K=32$}} & \textbf{$K=4$} & \textbf{$K=32$}\\ 
\Xhline{3\arrayrulewidth}

    ① & \multicolumn{1}{c}{2} & \multicolumn{1}{c|}{\multirow{3}{*}{0.25}}
    & \multicolumn{1}{r}{74.93} & \multicolumn{1}{r|}{80.37}
    & \multicolumn{1}{r}{72.55} & \multicolumn{1}{r}{77.73} \\

    ② & \multicolumn{1}{c}{4} & ~
    & \multicolumn{1}{r}{\textbf{75.09}} & \multicolumn{1}{r|}{\textbf{81.48}}
    & \multicolumn{1}{r}{\textbf{72.64}} & \multicolumn{1}{r}{\textbf{77.87}} \\

    ③ & \multicolumn{1}{c}{8} & ~
    & \multicolumn{1}{r}{74.65} & \multicolumn{1}{r|}{81.32}
    & \multicolumn{1}{r}{72.11} & \multicolumn{1}{r}{77.43} \\

\hline

    ④ & \multicolumn{1}{c}{\multirow{2}{*}{4}} & \multicolumn{1}{c|}{0.125}
    & \multicolumn{1}{r}{74.99} & \multicolumn{1}{r|}{81.21}
    & \multicolumn{1}{r}{72.47} & \multicolumn{1}{r}{77.61} \\

    ⑤ & ~ & \multicolumn{1}{c|}{0.5}
    & \multicolumn{1}{r}{74.51} & \multicolumn{1}{r|}{80.78}
    & \multicolumn{1}{r}{71.46} & \multicolumn{1}{r}{77.34} \\

\hline

    ⑥ & \multicolumn{1}{c}{8} & \multicolumn{1}{c|}{0.125}
    & \multicolumn{1}{r}{74.82} & \multicolumn{1}{r|}{81.25}
    & \multicolumn{1}{r}{71.65} & \multicolumn{1}{r}{77.60} \\

\Xhline{3\arrayrulewidth}
\end{tabular}
\vspace{-0.17cm}
\end{table}
\begin{table}[t!]
\centering
\caption{Modality Loss Weights Comparison on MDR2.\\
Bold numbers indicate the best performance.}
\vspace{-0.17cm}
\label{tab:SSL-weights}
\begin{tabular}{c|cc|cccc}
\Xhline{3\arrayrulewidth}
    \multirow{3}{*}{\makecell{\textbf{Exp} \\ \textbf{ID}}}
    & \multicolumn{2}{c|}{\textbf{Loss Weights}} 
    & \multicolumn{4}{c}{\textbf{F1-Macro Score @1 (\%)}} \\
\cline{2-7} 

    & \multirow{2}{*}{\makecell{\textbf{Video} \\ ($w_{\rm V}$)}} 
    & \multirow{2}{*}{\makecell{\textbf{Sensor} \\ ($w_{\rm S}$)}} 
    & \multicolumn{2}{c|}{\textbf{Noun Task}} 
    & \multicolumn{2}{c}{\textbf{Verb Task}} \\
\cline{4-7} 
    & & & \textbf{$K=4$} & \multicolumn{1}{c|}{\textbf{$K=32$}} & \textbf{$K=4$} & \textbf{$K=32$} \\ 
\Xhline{3\arrayrulewidth}

    ① & 0.5 & 0.5        
    & 75.09 & \multicolumn{1}{r|}{\textbf{81.48}}
    & \textbf{72.64} & \textbf{77.87} \\
    
    ② & 0.3 & 0.7
    & 74.56 & \multicolumn{1}{r|}{80.69}
    & 72.32 & 76.99 \\

    ③ & 0.7 & 0.3
    & \textbf{75.23} & \multicolumn{1}{r|}{81.18}
    & 71.60 & 76.59 \\

\Xhline{3\arrayrulewidth}
\end{tabular}
\vspace{-0.3cm}
\end{table}

\subsubsection{Input Temporal Sparsity Comparison}
We further evaluate how the temporal sparsity of pre-training inputs affects performance. 
As shown in Table~\ref{tab:SSL-inputs}, the selected sparsity configuration achieves the highest mean performance across all four evaluation settings, although several alternative configurations yield comparable results.

When the inputs are too sparse (Exp~①, ⑤), performance tends to decrease, especially for noun recognition with only two video frames. 
This suggests that overly sparse visual inputs may provide insufficient cues for effective pre-training. 
Conversely, denser inputs (Exp~③, ④, ⑥) also show slightly lower mean performance, likely because the reconstruction task becomes less challenging and provides weaker incentive for pre-training.

\subsubsection{Modality Loss Weights Comparison}
\label{subsec:abliation-weights}
MDR2's loss function consists of two reconstruction losses for different modalities; therefore, we also investigate different combinations of modality weights in the reconstruction loss of our MDR2 pre-training framework.

As shown in Table~\ref{tab:SSL-weights}, the equal-weight setting achieves the best overall performance across the four evaluation settings.
Although increasing the video weight slightly improves noun recognition under the K=4 setting, it degrades the other settings, especially verb recognition.
This result suggests that balancing the visual and sensor reconstruction losses is important for maintaining overall performance across tasks and input sparsity levels.

\section{Conclusion, Limitations, and Future Work}
In this paper, we proposed MDR2, a teacher-guided multimodal pre-training framework for visual-inertial egocentric human activity recognition.
MDR2 introduces dual modality-oriented reconstruction targets, using two unimodal teachers to supervise a shared multimodal representation. This design provides modality-specific reconstruction supervision, while introducing no additional inference-time overhead.
Experiments on EvIs-Kitchen demonstrate the effectiveness of MDR2 under both full-observation and sparse-input settings. Especially under sparse-input conditions, MDR2 improves the F1-macro scores by 9.20 and 6.17 percentage points over the multimodal baseline on noun and verb recognition, respectively.
In addition, the comparison with EviMAE shows that MDR2 achieves a favorable efficiency-performance trade-off, while the compatibility experiment on CMU-MMAC suggests that the proposed objective can be integrated into another video-IMU pre-training backbone.

This work still has several limitations. 
First, MDR2 adopts a teacher-guided multimodal representation learning paradigm, where unimodal teachers still need to be adapted using training-split labels.
Second, although the compatibility result suggests preliminary portability to another backbone, the generalizability of MDR2 across more diverse egocentric tasks and activity domains remains to be explored. 
Third, the teacher-guided design requires additional computational and memory resources during pre-training for unimodal teacher adaptation and dual reconstruction heads, which have not been systematically analyzed in this work.

Future work will address both methodological improvements and practical extensions.
At the methodological and theoretical levels, we will investigate how dual-target reconstruction affects the preservation of modality-specific and complementary information in the shared representation. 
This includes analyzing the influence of teacher quality and the relative weighting of the two reconstruction objectives, particularly under sparse visual observations.
We will also explore more efficient and less label-dependent teacher construction, including label-free or weakly supervised alternatives. For practical extensions, we will quantify and reduce the computational and memory costs of teacher adaptation and dual-target pre-training, and evaluate MDR2 across more diverse datasets, egocentric tasks, activity domains, and video-IMU backbones.


\bibliographystyle{IEEEtran}
\bibliography{IEEEabrv, reference}

    \vspace{-1.2cm}
\begin{IEEEbiography}[{\includegraphics[width=1in,height=1.25in,clip,keepaspectratio]{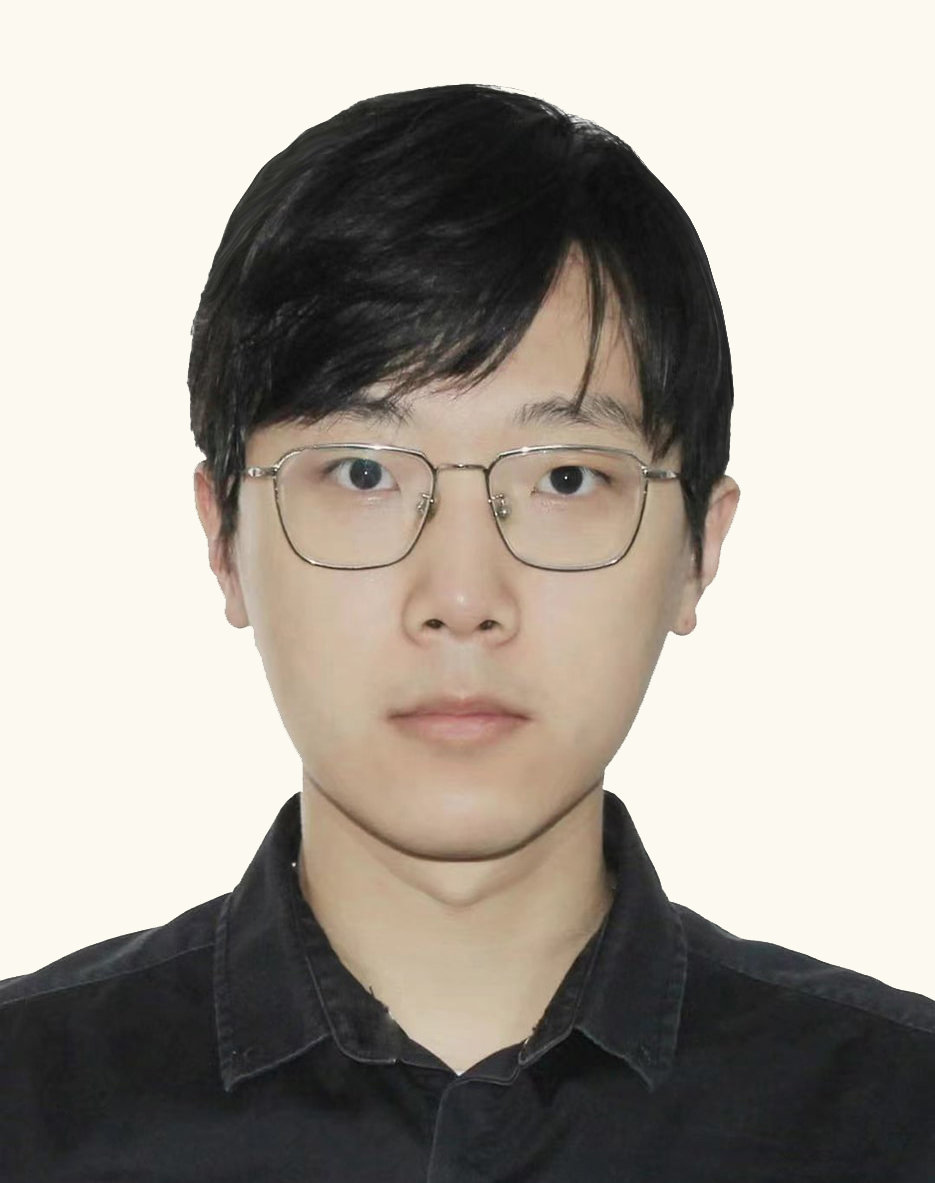}}]{Yuzhe Hao} received his B.S. degree in Information and Communication Engineering from the University of Electronic Science and Technology of China in 2019, and the M.S. degree in Computer Science from the Institute of Science Tokyo (formerly Tokyo Institute of Technology), Japan in 2021. He is currently a doctoral student in Artificial Intelligence at Institute of Science Tokyo. His research interests include action recognition, multimodal fusion, self-supervised learning and computer vision.
\end{IEEEbiography}
    \vspace{-1.2cm}

\begin{IEEEbiography}[{\includegraphics[width=1in,height=1.25in,clip,keepaspectratio]{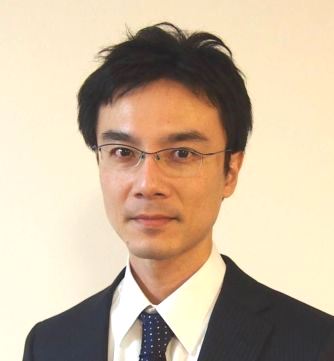}}]{Ikuro Sato} 
earned his Ph.D. in physics from the University of Maryland in 2005. Following a postdoctoral fellowship at Lawrence Berkeley National Laboratory (2005–2007), he has been working for Denso IT Laboratory since 2008. He currently serves as a Specially Appointed Professor at the Institute of Science Tokyo, a position he has held since 2020. His research interests include computer vision and machine learning. He is a member of IEEE.
\end{IEEEbiography}
    \vspace{-1.5cm}

\begin{IEEEbiography}[{\includegraphics[width=1in,height=1.25in,clip,keepaspectratio]{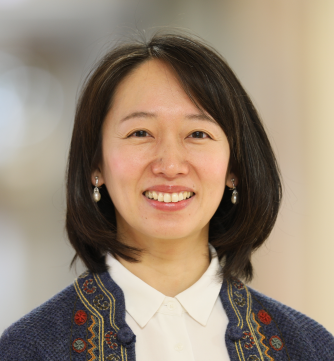}}]{Rei Kawakami} received her B.S., M.S., and Ph.D. degrees in information science and technology from the University of Tokyo in 2003, 2005, and 2008, respectively. She is currently an Associate Professor at Institute of Science Tokyo, Japan. Before that, she worked as a Senior Researcher at Denso IT Laboratory. Her research interests are in computer vision and image processing. She is a member of IEEE, ACM, IEICE, and IPSJ.
\end{IEEEbiography}
    \vspace{-1.5cm}

\begin{IEEEbiography}[{\includegraphics[width=1in,height=1.25in,clip,keepaspectratio]{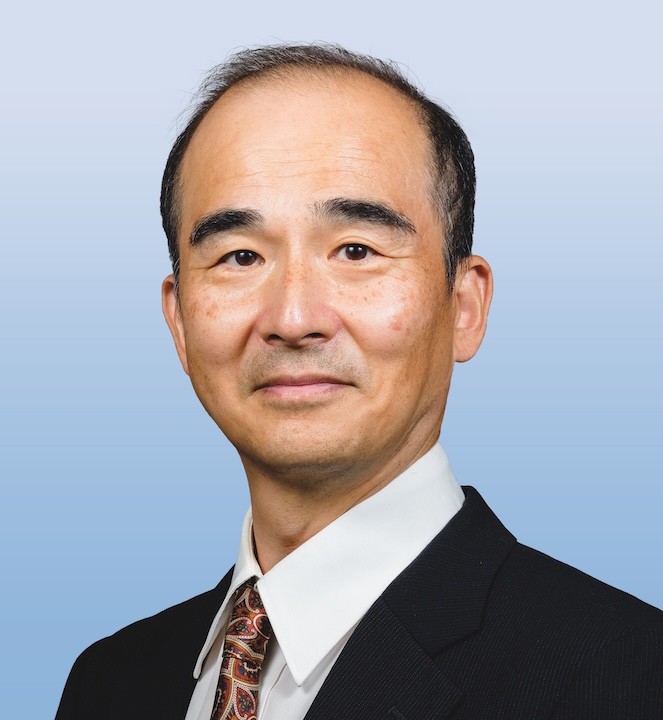}}]{Koichi Shinoda} received the BS and MS degrees from the University of Tokyo in 1987 and 1989, respectively, both in physics, and the DEng degree in computer science from the Institute of Science Tokyo in 2001. In 1989, he joined NEC Corporation, Japan, where he was involved in research on automatic speech recognition. From 1997 to 1998, he was a visiting scholar with Bell Labs, Lucent Technologies, Murray Hill, NJ. From October 2001 to March 2003, he was an associate professor at the University of Tokyo. 
He is currently a professor at the Institute of Science Tokyo. His research interests include speech recognition, video information retrieval, and machine learning. He was the chair of SIG-SLP (Spoken Language Processing) in Information Processing Society of Japan from 2014 to 2016 and the general co-chair of APSIPA 2021. He is a fellow of IEICE, and a senior member of IEEE and IPSJ.

\end{IEEEbiography}

\end{CJK}
\end{document}